\documentclass[10pt,twocolumn,letterpaper]{article}

\usepackage[pagenumbers]{cvpr} 

\usepackage[dvipsnames]{xcolor}

\definecolor{cvprblue}{rgb}{0.21,0.49,0.74}
\usepackage[pagebackref,breaklinks,colorlinks,allcolors=cvprblue]{hyperref}

\usepackage{multirow} 

\newcommand*{\affaddr}[1]{#1} 
\newcommand*{\affmark}[1][*]{\textsuperscript{#1}}
\newcommand*{\email}[1]{\texttt{#1}}

\usepackage{lipsum}

\title{DepthCrafter: Generating Consistent Long Depth Sequences \\ for Open-world Videos}

\author{
Wenbo Hu\affmark[1* \dag] \quad 
Xiangjun Gao\affmark[2*] \quad 
Xiaoyu Li\affmark[1* \dag] \quad 
Sijie Zhao\affmark[1] \quad \\
Xiaodong Cun\affmark[1] \quad 
Yong Zhang\affmark[1] \quad 
Long Quan\affmark[2] \quad 
Ying Shan\affmark[3, 1] \vspace{2mm}\\
\affaddr{\affmark[1]Tencent AI Lab}  \;
\affaddr{\affmark[2]The Hong Kong University of Science and Technology} \;
\affaddr{\affmark[3]ARC Lab, Tencent PCG} \\
\small\email{\url{https://depthcrafter.github.io}}
}

\begin{document}

\twocolumn[{
\maketitle
\begin{center}
    \captionsetup{type=figure}
    \vspace{-2em}
    \includegraphics[width=0.97\textwidth]{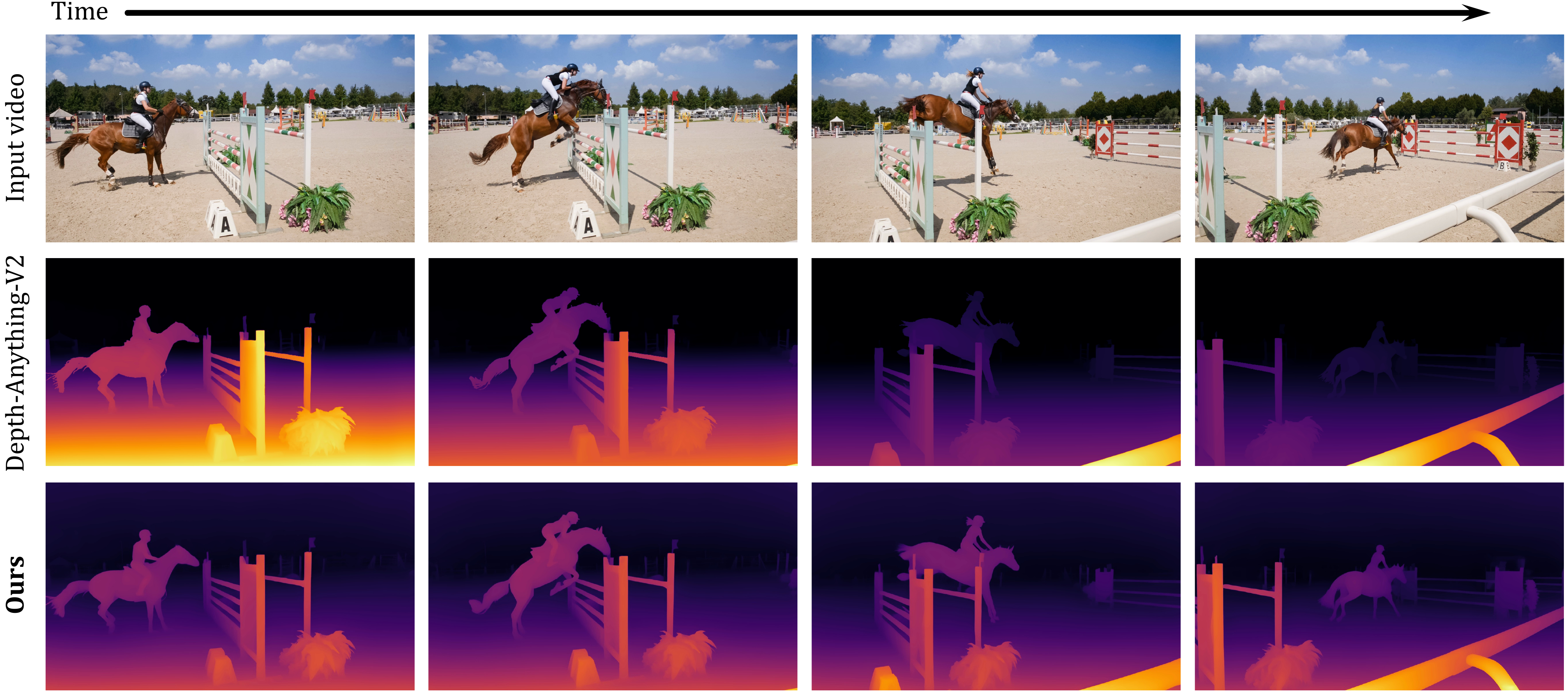}
    \vspace{-0.5em}
    \captionof{figure}{
    We innovate DepthCrafter, a novel video depth estimation approach, that can generate temporally consistent long depth sequences with fine-grained details for open-world videos, without requiring additional information such as camera poses or optical flow.
    }
    \vspace{-0.2em}
    \label{fig:teaser}
\end{center}
}]

\maketitle
\footnotetext[1]{~Joint first authors. 
\quad $^\dag$~Corresponding authors.
} 

\begin{abstract}
    \noindent 
    Estimating video depth in open-world scenarios is challenging due to the diversity of videos in appearance, content motion, camera movement, and length.
    We present DepthCrafter, an innovative method for generating temporally consistent long depth sequences with intricate details for open-world videos, without requiring any supplementary information such as camera poses or optical flow.
    The generalization ability to open-world videos is achieved by training the video-to-depth model from a pre-trained image-to-video diffusion model, through our meticulously designed three-stage training strategy.
    Our training approach enables the model to generate depth sequences with variable lengths at one time, up to 110 frames, and harvest both precise depth details and rich content diversity from realistic and synthetic datasets.
    We also propose an inference strategy that can process extremely long videos through segment-wise estimation and seamless stitching.
    Comprehensive evaluations on multiple datasets reveal that DepthCrafter achieves state-of-the-art performance in open-world video depth estimation under zero-shot settings.
    Furthermore, DepthCrafter facilitates various downstream applications, including depth-based visual effects and conditional video generation.
    %

\end{abstract}    
\section{Introduction}
\label{sec:intro}
Monocular depth estimation, serving as the bridge linking 2D observations and the 3D world, has been a long-standing fundamental problem in computer vision.
It plays a crucial role in a wide range of downstream applications, \eg, mixed reality, AI-generated content, autonomous driving, and robotics~\cite{Yu2022MonoSDF,hu2023Tri-MipRF,hong2022depth,hu-2021-bidirectional,liu20213d,dong2022towards,sun2021neuralrecon,yu2024viewcrafter}.
The inherent ambiguity makes it challenging, as the observed information from a single view is insufficient to determine the depth of a scene uniquely.

With recent advances in foundation models, we have witnessed significant progress in depth estimation from monocular images~\cite{oquab2023dinov2,darcet2023vitneedreg,depthanything,depth_anything_v2, marigold,fu2024geowizard,piccinelli2024unidepth,yin2023metric3d}.
However, all these methods are tailored for static images, without considering the temporal information in videos.
Temporal inconsistency, or flickering, would be observed when directly applying them to videos, as shown in Fig.~\ref{fig:teaser}.
Existing video depth estimation methods~\cite{Luo-VideoDepth-2020,zhang2021consistent,kopf2021rcvd,teed2018deepv2d,wang2023neural} typically try to optimize temporally consistent depth sequences from a pre-trained image depth model, with a given or learnable camera poses.
Their performance is sensitive to both the proportion of dynamic content and the quality of the camera poses. 
Yet, videos in the open world are diverse in content, motion, camera movement, and length, making these methods hard to perform well in practice.
Moreover, the required camera poses are usually non-trivial to obtain in open-world videos, particularly for long videos and videos with abundant dynamic content.

In this paper, we aim to generate temporally consistent long depth sequences with high-fidelity details for diverse open-world videos, without requiring any additional information.
Observing the strong capability of diffusion models in generating various types of videos~\cite{ho2022video,sora,blattmann2023stable,chen2023videocrafter1,chen2024videocrafter2,xing2023dynamicrafter,blattmann2023align,xing2024tooncrafter}, we propose \emph{DepthCrafter}, to leverage the video diffusion model for video depth estimation, while maintaining the generalization ability to open-world videos.
To train our DepthCrafter, a video-to-depth model, from a pre-trained image-to-video diffusion model, we compile paired video-depth datasets in two styles, \ie realistic and synthetic, since the realistic dataset provides rich content diversity and the synthetic dataset offers precise depth details.
On the aspect of temporal context, most existing video diffusion models can only produce a fixed and small number of frames at one time, \eg, 25 frames in Stable Video Diffusion (SVD)~\cite{blattmann2023stable}, which, however, is usually too short for open-world video depth estimation to accurately arrange depth distributions throughout the video.
To enable variable long temporal context and fuse the respective advantages of the two-styled datasets, we present a three-stage training strategy to progressively train certain layers of the diffusion model on different datasets with variable lengths.
By doing so, we can adapt the video diffusion model to generate depth sequences with variable lengths at one time, up to 110 frames, and harvest both the precise depth details and rich content diversity.
To further support extremely long videos, we tailor an inference strategy to process the video in overlapped segments and seamlessly stitch them together.

We extensively evaluate our DepthCrafter on diverse datasets, including indoor, outdoor, static, dynamic, realistic, and synthetic videos, under zero-shot settings.
Both qualitative and quantitative results demonstrate that our DepthCrafter achieves state-of-the-art performance in open-world video depth estimation, outperforming existing methods by a large margin.
Besides, we demonstrate that our DepthCrafter facilitates various downstream applications, including depth-based visual effects and conditional video generation.
Our contributions are summarized below:
\begin{itemize}
    \item We innovate DepthCrafter, a novel method to generate temporally consistent long depth sequences with fine-grained details for open-world videos, outperforming existing approaches by a large margin.
    
    \item We present a three-stage training strategy to enable generating depth sequences with a long and variable temporal context, up to 110 frames. It also allows us to harvest both the precise depth details and rich content diversity from synthetic and realistic datasets. 
    
    \item We design an inference strategy to segment-wisely process videos beyond 110 frames and seamlessly stitch them together, facilitating depth estimation for extremely long videos.
    
\end{itemize}

\section{Related Work}
\label{sec:related}

\noindent\textbf{Image depth estimation.}
Image depth estimation aims at predicting the depth map from a single image~\cite{eigen2014depth, fu2018deep, lee2019big, aich2021bidirectional, li2023depthformer, yang2021transformer, patil2022p3depth}.
However, the generalization ability to diverse open-world scenes is hindered by the limited training data.
To this end, MiDaS~\cite{ranftl2020towards} presented the affine-invariant depth representation, enabling mixed training datasets.
Depth-Anything (V2)~\cite{depthanything, depth_anything_v2} followed this idea and proposed to train the model on both labeled and large-scale unlabeled images, achieving good generalization ability.
Marigold~\cite{marigold} and follow-up works~\cite{fu2024geowizard, garcia2024fine, he2024lotus} leverage the diffusion priors to realize zero-shot transfer to unseen datasets.
Besides the affine-invariant depth, another stream of methods tried to estimate the absolute metric depth~\cite{bhat2023zoedepth, piccinelli2024unidepth, yin2023metric3d,hu2024metric3d,bochkovskii2024depth}.
All these methods are tailored for static images without considering the temporal consistency, while our work aims to generate temporally consistent long depth sequences for open-world videos.

\begin{figure*}[!t]
    \centering
    \vspace{-0.5em}
    \includegraphics[width=0.97\textwidth]{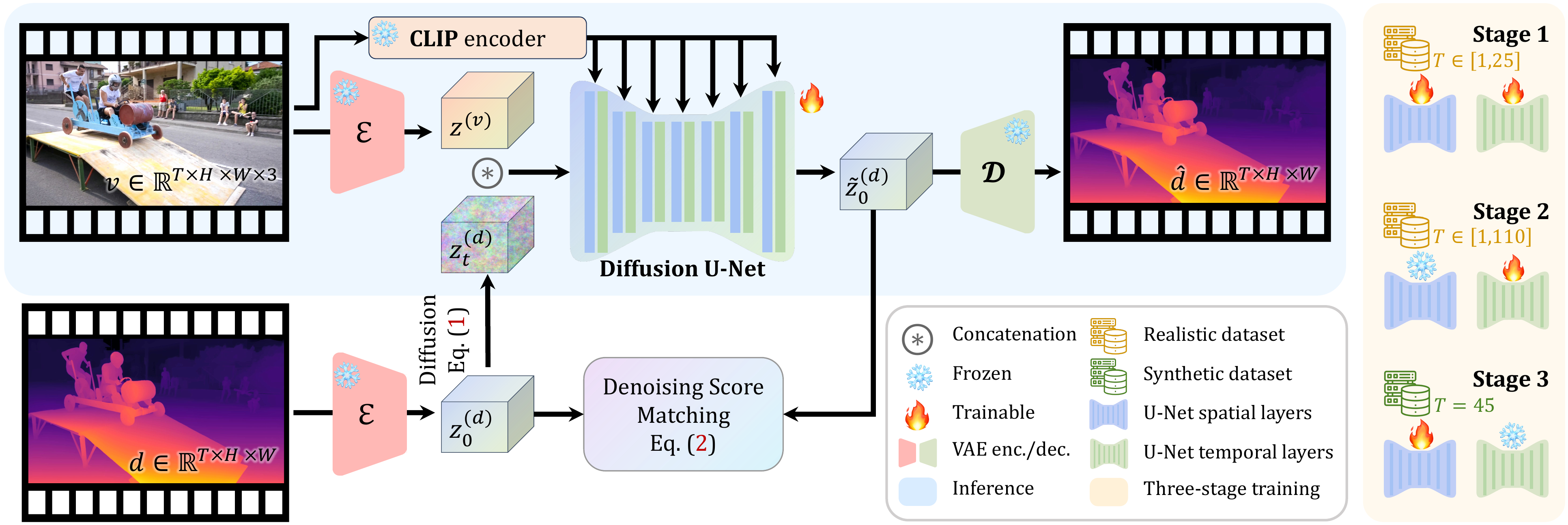}
    \vspace{-0.5em}
    \caption{
    Overview of our \emph{DepthCrafter}. It is a conditional diffusion model that models the distribution $p(\mathbf{d} \, | \, \mathbf{v})$ over the depth sequence $\mathbf{d}$ conditioned on the input video $\mathbf{v}$. 
    We train the model in three stages, where the spatial or temporal layers of the diffusion model are progressively learned on our compiled realistic or synthetic datasets with variable lengths $T$.
    During inference, given an open-world video, it can generate temporally consistent long depth sequences with fine-grained details for the entire video from initialized Gaussian noise, without requiring any supplementary information, such as camera poses or optical flow. 
    }
    \vspace{-1em}
    \label{fig:overview}
\end{figure*}

\vspace{0.3em}
\noindent\textbf{Video depth estimation.} 
Existing video depth estimation methods could be categorized into two classes: test-time optimization and feed-forward prediction.
Test-time optimization methods~\cite{Luo-VideoDepth-2020,kopf2021rcvd,zhang2021consistent,chen2019self} involve an optimization procedure for each video during inference, which typically requires camera poses or optical flow.
Their results are usually consistent, but the requirement of camera poses or optical flow would limit their applicability to open-world videos.
Feed-forward prediction methods directly predict depth sequences from videos~\cite{li2023temporally, wang2019web, teed2018deepv2d, wang2023neural, zhang2019exploiting, wang2022less, yasarla2023mamo, yasarla2024futuredepth}, \eg, DeepV2D~\cite{teed2018deepv2d} combines camera motion estimation with depth estimation, MAMO~\cite{yasarla2023mamo} leverages memory attention, and NVDS~\cite{wang2023neural} introduces a plug-and-play stabilization network.
However, due to the limited video depth training data, these methods often fail to address the in-the-wild videos with diverse content. 
By leveraging the pre-trained video diffusion model and our designed three-stage training strategy, our method demonstrates a powerful ability for open-world video depth estimation.

\vspace{0.3em}
\noindent\textbf{Video diffusion models.}
Diffusion models~\citep{sohl2015deep, ho2020denoising} have shown remarkable video generation ability~\cite{ho2022video, sora, blattmann2023stable, chen2023videocrafter1, chen2024videocrafter2, xing2023dynamicrafter, blattmann2023align, xing2024tooncrafter}.
Among these methods, VDM~\cite{ho2022video} presents the first results on video generation using diffusion models, Sora~\cite{sora} achieves impressive performance in this area, and SVD~\cite{blattmann2023stable} provides the popular open-source models for image-to-video generation. 
Trained on a well-curated video dataset, SVD can generate high-quality videos and is used as the model prior for various video-related tasks. 
In this paper, we leverage the video diffusion model for high-fidelity consistent video depth estimation, such that the generalization ability to open-world videos can be maintained.
Concurrent to our work, ChronoDepth~\cite{shao2024learning} also explores video depth estimation with video diffusion priors.
However, ChronoDepth only supports a short temporal context, \ie 10 frames, which is insufficient to accurately arrange depth distributions throughout the video.
In contrast, our method not only supports variable-length temporal context at one time, up to 110 frames, but also can estimate depth sequences for extremely long videos.

\section{Method}
\label{sec:method}

Given an open-world video, $\mathbf{v} \in \mathbb{R}^{T \times H \times W \times 3}$, our goal is to estimate temporally consistent depth sequences, ${\mathbf{d}} \in \mathbb{R}^{T \times H \times W}$, with fine-grained details.
%
%
Considering the diversity of open-world videos in appearance, content motion, camera movement, and length, the challenges to achieving our goal are threefold:
1.) a comprehensive understanding of video content for generalization ability;
2.) a long and variable temporal context to arrange the entire depth distributions accurately and keep temporal consistency;
and 3.) the ability to process extremely long videos.
As shown in Fig.~\ref{fig:overview}, we tackle these challenges by formulating the video depth estimation as a conditional diffusion generation problem $p(\mathbf{d} \, | \, \mathbf{v})$.
We train a video-to-depth model from a pre-trained image-to-video diffusion model through a meticulously designed three-stage training strategy with compiled paired video-depth datasets, and tailor an inference strategy to process extremely long videos through segment-wise estimation and seamless stitching.

\subsection{Preliminaries of Video Diffusion Models}
\label{sec:preliminaries}
Diffusion models~\cite{ho2020denoising,sohl2015deep} learn the data distribution $p(\mathbf{x})$ by a forward diffusion process to gradually noise the data to a target distribution, \eg the Gaussian distribution, and a reverse denoising process to iteratively recover the data from the noise by a learned denoiser.
In this paper, our study is conducted based on the stale video diffusion (SVD)~\cite{blattmann2023stable}, which is a famous open-source video diffusion model.
SVD adopts the EDM~\cite{karras2022elucidating} diffusion framework.
The diffusion process is achieved by adding \iid $\sigma_t^2$-variance Gaussian noise to the data $\mathbf{x}_0 \sim p(\mathbf{x})$:
\begin{equation}
    \mathbf{x}_t = \mathbf{x}_0 + \sigma_t^2 \mathbf{\epsilon}, \quad \mathbf{\epsilon} \sim \mathcal{N}\big{(}\mathbf{0}, \mathbf{I}\big{)},
    \label{eq:diffusion}
\end{equation}
where $\mathbf{x}_t \sim p(\mathbf{x};\sigma_t)$ is the data with noise level $\sigma_t$.
When $\sigma_t$ is large enough ($\sigma_\text{max}$),
the distribution would be indistinguishable from the Gaussian distribution.
%
Based on this fact, the diffusion model starts from a high-variance Gaussian noise $\mathbf{\epsilon} \sim \mathcal{N}\big{(}\mathbf{0}, \sigma_{\text{max}}^{2} \mathbf{I}\big{)}$ and gradually denoises it towards $\sigma_0 = 0$ to generate the data.
The denoiser $D_{\theta}$ is a learnable function that tries to predict the clean data, \ie $\tilde{\mathbf{x}}_0 = D_{\theta} \big{(} \mathbf{x}_t; \sigma_t \big{)}$.
Its training objective is the denoising score matching:
\begin{equation}
    \mathbb{E}_{
        \mathbf{x}_t \sim p(\mathbf{x};\sigma_t), \sigma_t \sim p(\sigma)
        } 
        \bigg{[} \lambda_{\sigma_t} \bigg{ \Vert} D_{\theta} \big{(} \mathbf{x}_t; \sigma_t; \mathbf{c} \big{)} - \mathbf{x}_0 \bigg{ \Vert}_2^2 \bigg{]},
    \label{eq:denoising_loss}
\end{equation}
where $p(\sigma)$ is the noise level distribution during training, $\mathbf{c}$ denotes the conditioning information, and $\lambda_{\sigma_t}$ is the weight for the denoising loss at time $t$.
To promote the learning, EDM adopts the preconditioning strategy~\cite{karras2022elucidating,salimans2022progressive}, to parameterize the denoiser $D_{\theta}$ as:
\begin{equation}
    \begin{split}
        D_{\theta} &\big{(} \mathbf{x}_t; \sigma_t; \mathbf{c} \big{)} = \\
        &c_\text{skip}(\sigma_t)  \mathbf{x}_t + 
        c_\text{out}(\sigma_t)  
        F_{\theta} \big{(} 
            c_\text{in}  \mathbf{x}_t;  c_\text{noise}(\sigma_t); \mathbf{c}
        \big{)},
    \end{split}
    \label{eq:denoiser}
\end{equation}
where $F_{\theta}$ is implemented as a learnable U-Net~\cite{ronneberger2015u}, and $c_\text{in}$, $c_\text{out}$, $c_\text{skip}$, and $c_\text{noise}$ are preconditioning functions.

\subsection{Formulation with Diffusion Models}
\label{sec:formulation}

\noindent\textbf{Latent space transformation.}
To generate high-resolution depth sequences without sacrificing computational efficiency, we adopt the framework of Latent Diffusion Models (LDMs)~\cite{rombach2022high} that perform in a low-dimensional latent space.
The transformation between the latent and data spaces is achieved by a Variational Autoencoder (VAE)~\cite{kingma2013auto}, which was originally designed for encoding and decoding video frames in SVD~\cite{blattmann2023stable}.
Fortunately, we found it can be directly used for depth sequences with only a negligible reconstruction error, which is similar to the observation in Marigold~\cite{marigold} for image depth estimation.
As shown in Fig.~\ref{fig:overview}, the latent space transformation is formulated as:
\begin{equation}
    \mathbf{z}^{(\mathbf{x})} = \mathcal{E}(\mathbf{x}), \quad \hat{\mathbf{x}} = \mathcal{D} \big{(} \mathbf{z}^{(\mathbf{x})} \big{)},
    \label{eq:latent_transformation}
\end{equation}
where $\mathbf{x}$ is either the video $\mathbf{v}$ or the depth sequence $\mathbf{d}$, $\mathbf{z}^{(\mathbf{x})}$ is the latent representation of the data, $\hat{\mathbf{x}}$ is the reconstructed data, $\mathcal{E}$ and $\mathcal{D}$ are encoder and decoder of the VAE, respectively.
For the depth sequence, we replicate it three times to meet the 3-channel input format of the encoder in VAE and average the three channels of the decoder output to obtain the final latent of the depth sequence.
Following the practice in image depth estimation~\cite{ranftl2020towards,ranftl2021vision,depthanything,depth_anything_v2,marigold}, we adopt the relative depth, \ie the affine-invariant depth, which is normalized to $[0, 1]$.
But differently, our predicted depth sequence shares the same scale and shift across frames, rather than a per-frame normalization, which is crucial for maintaining temporal consistency.

\noindent\textbf{Conditioning on the video.}
SVD is an image-to-video diffusion model that generates videos conditioned on a single image.
The conditional image is fed into the U-Net in two ways, \ie, concatenating its latent to the input latent, and injecting its CLIP~\cite{radford2021learning} embedding to the intermediate features via cross-attention.
Yet, our DepthCrafter involves the generation of depth sequences conditioned on video frames in a frame-to-frame fashion.
Therefore, we adapt the conditioning mechanism to meet our video-to-depth generation task.
As shown in Fig.~\ref{fig:overview}, given the encoded latent of depth sequence $\mathbf{z}^{(\mathbf{d})}$ and video frames $\mathbf{z}^{(\mathbf{v})}$ from Eq.~\eqref{eq:latent_transformation}, we concatenate the video latent to the input noisy depth latent frame-wisely, rather than only the first frame, to condition the denoiser for generating the depth sequence.
For high-level semantic information, we embed the video frames using CLIP and then inject the embeddings in a frame-to-frame manner to the denoiser via cross-attention.
Compared to the original conditioning mechanism, our adapted conditioning provides comprehensive information from the video frames to the denoiser, which guarantees the alignment between the generated depth sequences and the video content.

\subsection{Training Strategy}
\label{sec:training}
To train our DepthCrafter, we need a large amount of high-quality paired video-depth sequences.
Although there are several video depth datasets available, \eg, KITTI~\cite{Geiger2013IJRR}, Scannet~\cite{dai2017scannet}, VDW~\cite{Wang_2023_ICCV}, DynamicReplica~\cite{karaev2023dynamicstereo}, and MatrixCity~\cite{li2023matrixcity}, they are either lacking high-quality depth annotations or restricted to a specific domain, \eg, driving scenes, indoor scenes, or synthetic scenes.

\noindent\textbf{Dataset construction.}
To this end, we compiled paired datasets of two styles, \ie realistic and synthetic, where the realistic dataset is large-scale and diverse, and the synthetic dataset is miniature but fine-grained and accurate.
The realistic dataset is constructed from a large number of binocular videos with a wide range of scene and motion diversity.
We cut the videos according to scene changes, and apply the state-of-the-art video stereo matching method, \eg, BiDAStereo~\cite{jing2024matchstereovideos}, to generate temporally consistent depth sequences.
Finally, we obtained $\sim$200K paired video-depth sequences with the length of $50-200$ frames.
The synthetic dataset is a combination of the DynamicReplica~\cite{karaev2023dynamicstereo} and MatrixCity~\cite{li2023matrixcity} datasets, which contains $\sim$3K fine-grained depth annotations with a length of 150 frames.

\noindent\textbf{Challenges of variable long temporal context.}
Different from image depth estimation which can determine the distribution of relative depth from a single frame, the video depth estimation requires a long temporal context to arrange the depth distributions accurately for the entire video and keep the temporal consistency.
Besides, the model should support variable-length estimation as the length of open-world videos may vary significantly.
However, existing open-source video diffusion models can only generate a fixed small number of frames at a time, \eg, 25 frames in SVD~\cite{blattmann2023stable}.
It is non-trivial to adapt the pre-trained model to meet this requirement, as directly fine-tuning it with long sequences is memory-consuming, for example, modern GPUs with 40GB memory can only support the training of a 25-frame sequence in SVD.

\noindent\textbf{Three-stage training.}
Considering both the two-style paired datasets and the long temporal context requirement, we design a three-stage training strategy to harvest the variety of video content, the precise depth details, as well as the support for long and variable sequences.
As shown in Fig.~\ref{fig:overview}, we train our DepthCrafter from the pre-trained SVD in three stages.
We first train it on our large realistic dataset to adapt the model to the video-to-depth generation task.
The sequence length in this stage is randomly sampled from $[1, 25]$ frames, such that the model can learn to generate depth sequences with variable lengths.
In the second stage, we only fine-tune the temporal layers of the model, still on our large realistic dataset, but with the sequence length randomly sampled from $[1, 110]$ frames.
The reason why we only fine-tune the temporal layers is that the temporal layers are more sensitive to the sequence length while the spatial layers are already adapted to the video-to-depth generation task in the first stage, and doing so significantly reduces memory consumption compared to fine-tuning the full model.
The long temporal context in this stage enables the model to precisely arrange the entire depth distributions for long and variable sequences.
In the third stage, we fine-tune the spatial layers of the model on our small synthetic dataset, with a fixed sequence length of 45 frames since the model has already learned to generate depth sequences with variable lengths in the first two stages and tuning the spatial layers would not affect the temporal context.
%
As the depth annotations in the synthetic dataset are more accurate and fine-grained, the model can learn more precise depth details in this stage.
The three-stage training strategy makes our DepthCrafter capable of generating high-quality depth sequences for open-world videos with variable lengths.

\begin{figure}[!t]
    \centering
    \includegraphics[width=1.0\columnwidth]{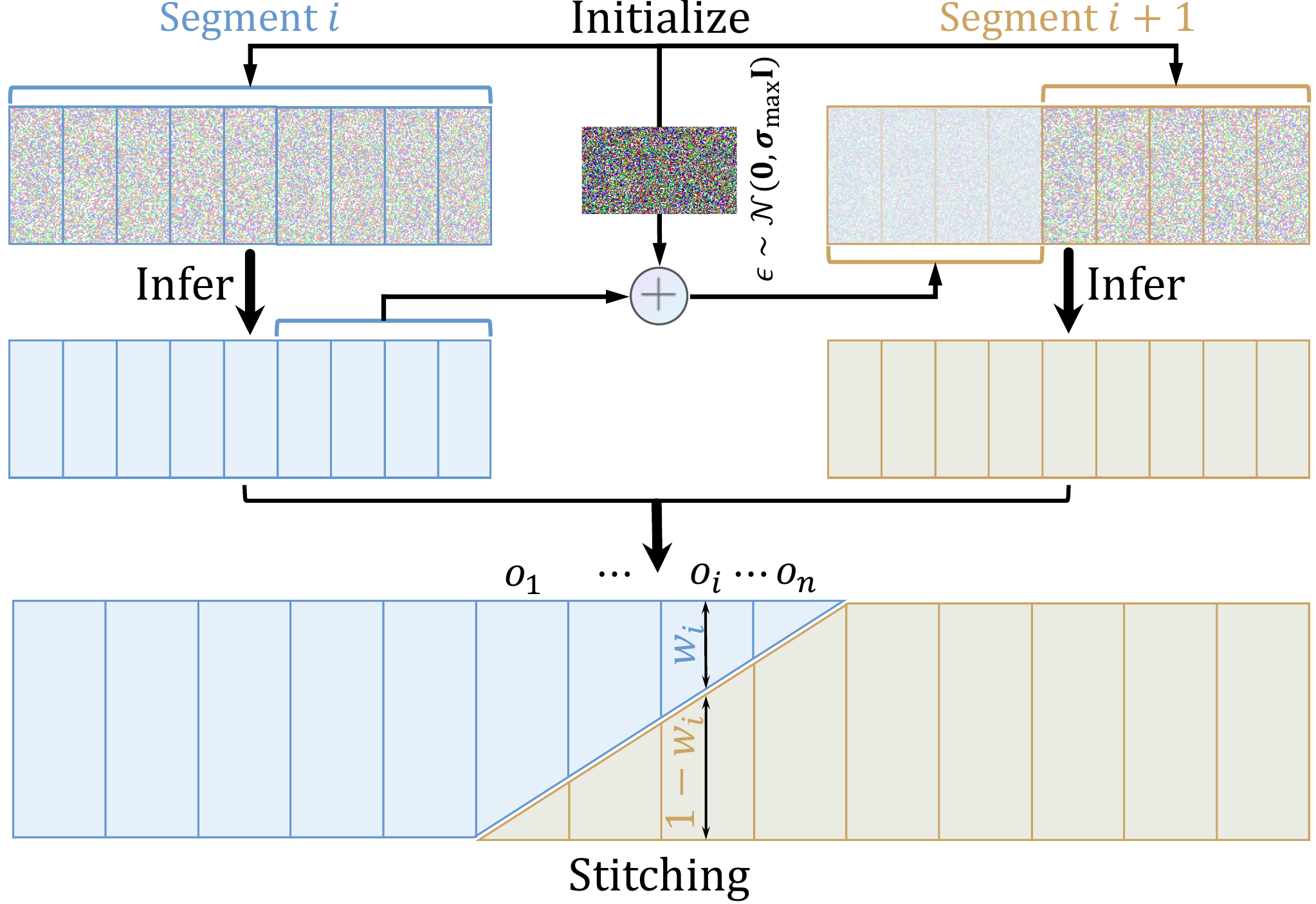}
    \vspace{-1.5em}
    \caption{
    Inference for extremely long videos. We divide the video into overlapped segments and estimate the depth sequence for each segment with a noise initialization strategy to anchor the scale and shift of depth distributions. These depth segments are then seamlessly stitched together with a latent interpolation strategy to form the entire depth sequence. The overlapped frames and their interpolation weights are denoted as ${o}_i$, ${w}_i$ and $1-{w}_{i}$, respectively.
    }
    \vspace{-1em}
    \label{fig:inference}
\end{figure}

\subsection{Inference for Extremely Long Videos}
\label{sec:inference}
Although the model can estimate depth sequences up to the length of 110 frames after training, it is still far from long enough for open-world videos, which can even contain hundreds or thousands of frames.
To this end, we design an inference strategy to infer extremely long depth sequences in a segment-wise manner and seamlessly stitch them together to form the entire depth sequence.
As shown in Fig.~\ref{fig:inference}, we first divide the video into overlapped segments, whose lengths are up to 110 frames.
Then we estimate the depth sequences for each segment.
Rather than purely initializing the input latent with Gaussian noise $\mathbf{\epsilon} \sim \mathcal{N}\big{(}\mathbf{0}, \mathbf{\sigma}_\text{max}^{2} \mathbf{I}\big{)}$, we initialize the latent of the overlapped frames by adding noise to the denoised latent from the previous segment, to anchor the scale and shift of the depth distributions.
Finally, to further ensure the temporal smoothness across segments, we craft a mortise-and-tenon style latent interpolation strategy to stitch consecutive segments together, inspired by~\cite{zhang2024mimicmotion}.
Specifically, we interpolate the latent of the overlapped frames ${o}_i$ from the two segments with the interpolation weights ${w}_i$ and $1-{w}_{i}$, respectively, where ${w}_i$ is linearly decreased from 1 to 0.
The final estimated depth sequence is obtained by decoding the stitched latent segments with the decoder $\mathcal{D}$ in the VAE.
With the training and inference strategies, our DepthCrafter can generate temporally consistent long depth sequences for open-world videos.

\begin{figure*}[!t]
    \centering
    \includegraphics[width=1.0\linewidth]{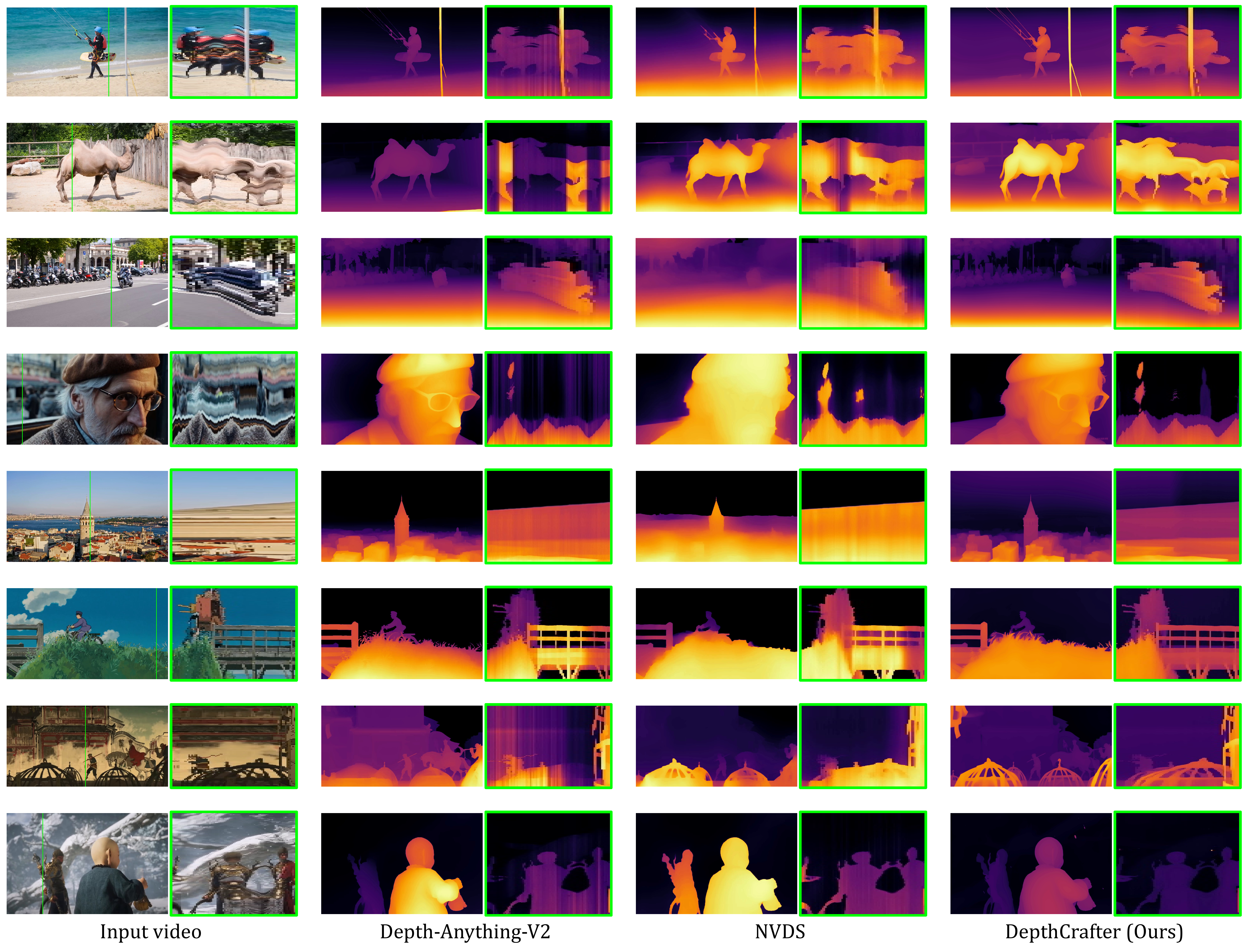}
    \vspace{-2em}
    \caption{
    Qualitative comparison for open-world video depth estimation. We compare with the representative single-image depth estimation method Depth-Anything-V2~\cite{depth_anything_v2} and the video depth estimation method NVDS~\cite{wang2023neural}. For better visualizing the temporal quality, we show the temporal profiles of each result in green boxes, by slicing the depth values along the time axis at the green line positions.
    }
    \vspace{-1.5em}
    \label{fig:qualitative}
\end{figure*}

\section{Experiments}
\label{sec:experiments}

\subsection{Implementation}
We implemented our DepthCrafter based on SVD~\cite{blattmann2023stable}, using the diffusers~\cite{von-platen-etal-2022-diffusers} library. 
We train our model at the resolution of $320 \times 640$ for efficiency, but we can estimate depth sequences at any resolution, \eg, $576 \times 1024$, during inference.
We use the Adam optimizer~\cite{kingma2014adam} with a learning rate of $1\times10^{-5}$ and a batch size of 8.
The number of iterations in the three stages of training is $80K$, $40K$, and $10K$, respectively.
We employed eight NVIDIA A100 GPUs for training, with a total training time of about five days.
%
%
Except for specific explanations, the number of denoising steps is set to \textbf{5} during inference.
Please refer to the supplementary material for more implementation details.

\subsection{Evaluation}

\noindent\textbf{Evaluation datasets.} 
We evaluate our model on four video datasets, a single-image dataset, as well as the DAVIS dataset~\cite{Perazzi2016} and in-the-wild videos for qualitative results.
The evaluations are conducted under the \emph{zero-shot} setting, where the testing datasets cover a variety of scenes, including synthetic and realistic scenes, indoor and outdoor scenes, and static and dynamic scenes, to evaluate the generalization ability of our model across various open-world scenarios.
{Sintel}~\cite{Sintel} is a synthetic dataset with precise depth labels, featuring dynamic scenes with diverse content and camera motion.
It contains 23 sequences with a length of around 50 frames.
{ScanNet v2}~\cite{dai2017scannet} is an indoor dataset with depth maps obtained from a Kinect sensor.
We extracted 90 frames at a rate of 15 frames per second from each sequence of the test set, which includes 100 RGB-D video sequences of various scenes.
Since ScanNet v2 contains only static indoor scenes, we further introduced 5 dynamic indoor RGB-D videos with a length of 110 frames each from the {Bonn}~\cite{palazzolo2019iros} dataset to better evaluate the performance of our model on dynamic scenes.
{KITTI}~\cite{Geiger2013IJRR} is a street-scene outdoor dataset for autonomous driving, with sparse metric depths captured by a LiDAR sensor.
We adopted the validation set, which includes 13 scenes, and extracted 13 videos from it with a length of 110 frames each.
Besides, we also evaluated our model for single-image depth estimation on the {NYU-v2}~\cite{silberman2012indoor} test set, which contains 654 images.

\begin{table*}[!t]
    \centering
    \caption{{Zero-shot depth estimation results}. We compare with the best single-image depth estimation model, Marigold~\cite{marigold}, and Depth Anything (V2)~\cite{depthanything,depth_anything_v2}, as well as the representative video depth estimation models, NVDS~\cite{wang2023neural} and ChronoDepth~\cite{shao2024learning}. 
    \textbf{Best} and \underline{second best} results are highlighted.
    }
    \vspace{-0.5em}
    \resizebox{\linewidth}{!}{
        \begin{tabular}{lcccccccccc}
        \toprule
        \multirow{2}{*}{Method} & \multicolumn{2}{c}{Sintel ($\sim$50 frames)} & \multicolumn{2}{c}{Scannet (90 frames)} & \multicolumn{2}{c}{KITTI (110 frames)} & \multicolumn{2}{c}{Bonn (110 frames)} & \multicolumn{2}{c}{NYU-v2 (1 frame)} \\
    
        \cmidrule(lr){2-3}\cmidrule(lr){4-5}\cmidrule(lr){6-7}\cmidrule(lr){8-9}\cmidrule(lr){10-11}
        
        ~ & AbsRel $\downarrow$ & $\delta_1$ $\uparrow$ & AbsRel $\downarrow$ & $\delta_1$ $\uparrow$ & AbsRel $\downarrow$ & $\delta_1$ $\uparrow$ & AbsRel $\downarrow$ & $\delta_1$ $\uparrow$ & AbsRel $\downarrow$ & $\delta_1$ $\uparrow$ \\
        
        \midrule
        NVDS~\cite{wang2023neural} & 
        0.408 & 0.483 & 
        0.187 & {0.677} &      
        0.253 & 0.588 &    
        0.167 & 0.766 & 
        0.151 & 0.780  \\
    
        ChronoDepth~\cite{shao2024learning} & 
        0.587 & 0.486 & 
        0.159 & 0.783 &
        0.167 & 0.759 & 
        0.100 & 0.911 &
        0.073 & 0.941  \\

        Marigold~\cite{marigold} & 
        0.532 & 0.515 &  
        0.166 & 0.769 & 
        0.149 & 0.796 & 
        0.091 & 0.931 & 
        0.070 & 0.946  \\
        {{Depth-Anything}~\cite{depthanything}} & 
        \underline{0.325} & \underline{0.564} & 
        \underline{0.130} & \underline{0.838} & 
        {0.142} & {0.803} & 
        \underline{0.078} & \underline{0.939} & 
        \textbf{0.042} & \textbf{0.981}\\
        {{Depth-Anything-V2}~\cite{depth_anything_v2}} & 
        0.367 & 0.554 & 
        0.135 & 0.822 & 
        \underline{0.140} & \underline{0.804} & 
        0.106 & 0.921 & 
        \underline{0.043} & \underline{0.978}\\

        \midrule
        {DepthCrafter (Ours)} & 
        \textbf{0.270} & \textbf{0.697} & 
        \textbf{0.123} & \textbf{0.856} & 
        \textbf{0.104} & \textbf{0.896} & 
        \textbf{0.071} & \textbf{0.972} &
        {0.072} & {0.948}\\
        
        \bottomrule
        \end{tabular}
    }    
    \vspace{-1em}
    \label{tab:zeroshot_mde}
\end{table*}

\noindent\textbf{Evaluation metrics.}
Following practice in affine-invariant depth estimation~\cite{dong2022towards,depthanything,depth_anything_v2,marigold}, we align the estimated depth maps with the ground truth using a scale and shift, and calculate two metrics: AbsRel $\downarrow$ (absolute relative error: $|\hat{\mathbf{d}}-\mathbf{d}| / \mathbf{d}$) and $\delta_1\uparrow$ (percentage of $\max(\mathbf{d}/\mathbf{\hat{d}}, \mathbf{\hat{d}}/\mathbf{d}) < 1.25$).
Different from previous methods that optimize the scale and shift individually for each frame, we optimize \emph{a shared scale and shift across the entire video}, which is more challenging, but necessary, for video depth estimation to ensure temporal consistency.

\noindent\textbf{Quantitative results.}
We compare our DepthCrafter with the representative methods for both single-image and video depth estimation, \ie Marigold~\cite{marigold}, Depth-Anything~\cite{depthanything}, Depth-Anything-V2~\cite{depth_anything_v2}, NVDS~\cite{wang2023neural}, and ChronoDepth~\cite{shao2024learning}.
For Depth-Anything (V2), we use the large model variant for its best performance, while for Marigold we adopt the LCM version with an ensemble size of 5.
As shown in Tab.~\ref{tab:zeroshot_mde}, our DepthCrafter achieves state-of-the-art performance in all four video datasets, thanks to the powerful open-world video undederstanding capability of the video diffusion models and the three-stage training strategy that leverages both realistic and synthetic datasets.
For Sintel and KITTI, characterized by significant camera motion and fast-moving objects, our DepthCrafter outperforms the strong Depth-Anything (V2) model tremendously in terms of both the AbsRel and $\delta_1$ metrics, \eg $25.7\%=(0.140-0.104)/0.140$ improvement in AbsRel on KITTI.
For indoor datasets like Scannet and Bonn, featuring minimal camera motion and roughly the same room scales, Depth-Anything has exhibited good performance. 
Nevertheless, we still have performance enhancements over it, \eg $5.4\%=(0.130-0.123)/0.130$ improvement in AbsRel on Scannet.
Note that the sequence length of these datasets varies from 50 to 110 frames, and our model can generalize well across different video lengths.

\begin{figure}[!t]
    \centering
    \includegraphics[width=1.0\columnwidth]{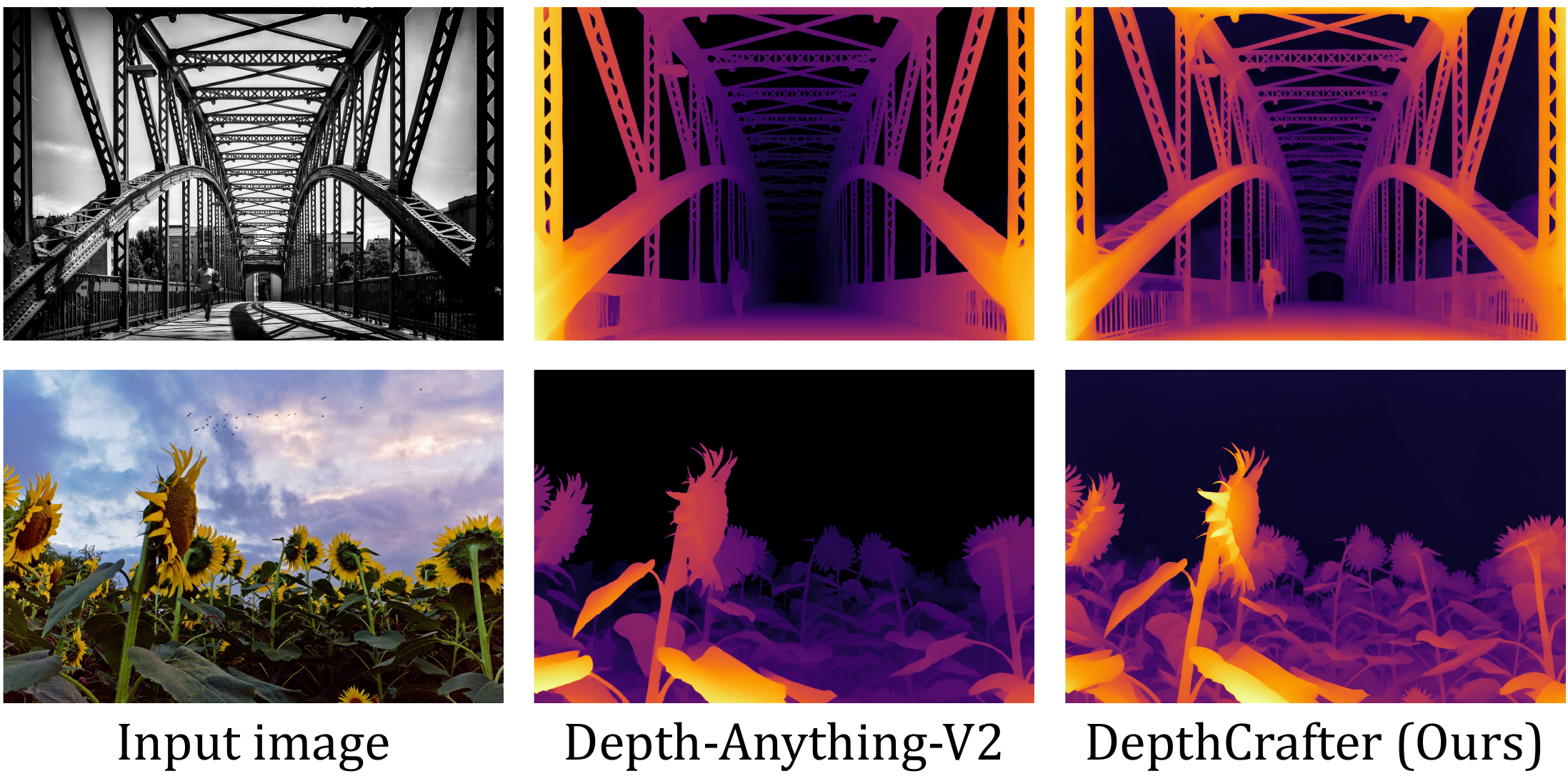}
    \vspace{-1.5em}
    \caption{
    Single-image depth estimation. We compare with the representative single-image depth estimation method Depth-Anything-V2~\cite{depth_anything_v2}.
    }
    \vspace{-2em}
    \label{fig:image}
\end{figure}

\noindent\textbf{Qualitative results.}
To further demonstrate the effectiveness of our model, we present the qualitative results from the DAVIS dataset~\cite{Perazzi2016}, Sora generated videos~\cite{sora}, and open-world videos, including human actions, animals, architectures, cartoons, and games, where the sequence length varies from 90 to 195 frames.
As shown in Fig.~\ref{fig:qualitative}, we show the temporal profiles of the estimated depth sequences in the green line position by slicing the depth values along the time axis, to better visualize the temporal consistency of the estimated depth sequences, following the practice in~\cite{wang2023neural,hu-2020-mononizing}.
We can observe that our DepthCrafter can produce temporally consistent depth sequences with fine-grained details across various open-world videos, while both NVDS and Depth-Anything exhibit zigzag artifacts in the temporal profiles, indicating the flickering artifacts in the estimated depth sequences.
These results demonstrate the effectiveness of our DepthCrafter in generating temporally consistent long depth sequences with high-fidelity details for open-world videos.

\noindent\textbf{Single-image depth estimation.}
Although our model is designed for video depth estimation, it can also perform single-image depth estimation, as our DepthCrafter can estimate video depth of any length.
As shown in Tab.~\ref{tab:zeroshot_mde}, our DepthCrafter achieves competitive performance in single-image depth estimation on the NYU-v2 dataset.
Since the depth labels in the NYU-v2 dataset are sparse and noisy, we also provide the qualitative results in Fig.~\ref{fig:image} to demonstrate the effectiveness of our model in estimating depth maps from static images.
We can observe that our DepthCrafter can even produce more detailed depth maps than Depth-Anything-V2, which is the existing state-of-the-art single-image depth estimation model.
These results demonstrate the ability of our DepthCrafter for processing both video and single-image depth estimation tasks.

\begin{table}[t]
    \small
        \centering
        \caption{Ablation study on the Sintel dataset. We evaluate the performance of our model at the end of different stages of training.}
        \vspace{-0.5em}
        \setlength\tabcolsep{5mm}
        \resizebox{\linewidth}{!}{
        \begin{tabular}{lccc}
            \toprule
            & stage 1 & stage 2 & stage 3 \\
            \midrule
            AbsRel $\downarrow$& {0.322} & \underline{0.316} & \textbf{0.270} \\
            $\delta_1$ $\uparrow$& 0.626 & \underline{0.675} & \textbf{0.697} \\
            \bottomrule
        \end{tabular}
        }
        \vspace{-1.5em}
        \label{tab:ablation}
\end{table}

\subsection{Ablation Studies}

\noindent\textbf{Effectiveness of the three-stage training strategy.}
We first ablate the effectiveness of the three-stage training strategy by evaluating the performance of our model at the end of each stage on the Sintel dataset~\cite{Sintel}, since it contains precise depth annotations on dynamic scenes.
From Tab.~\ref{tab:ablation}, we can observe that the performance of our model consistently improves as the number of training stages increases, indicating the effectiveness of the three-stage training strategy.
More ablation results on other datasets are available in the supplementary material.

\begin{figure}[!t]
    \centering
    \includegraphics[width=1.0\columnwidth]{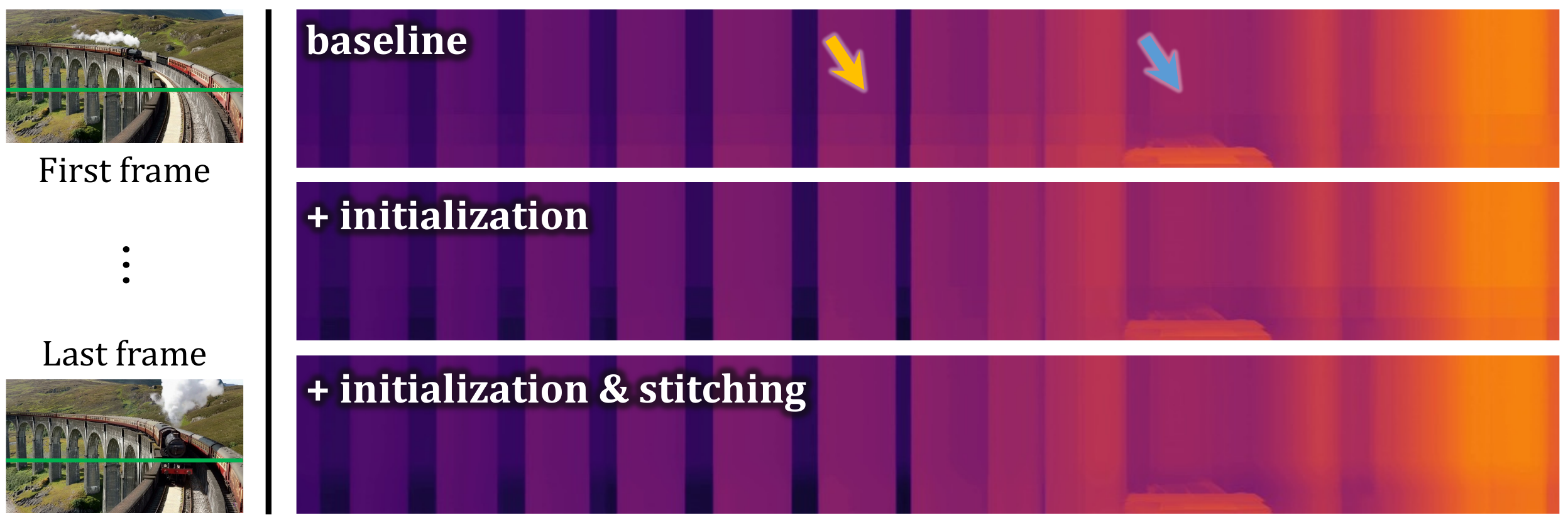}
    \vspace{-1.5em}
    \caption{
    Ablation studies on the effectiveness of the inference strategy. We profile the estimated depth sequences of different variants on the green line position. The yellow and blue arrows point to the static and dynamic regions, respectively.
    }
    \vspace{-1.5em}
    \label{fig:ablation}
\end{figure}

\noindent\textbf{Effectiveness of the inference strategy.}
To ablate the effectiveness of our inference strategy components, we consider these variants:
\textbf{baseline}, which independently infers each segment and directly averages the overlapped frames; 
\textbf{+ initialization}, which contains the same initialization of overlapped latents as our method, but without the stitching process;
\textbf{+ initialization \& stitching}, which is our full method.
We visually compare the temporal profiles of the estimated depth sequences of these variants in Fig.~\ref{fig:ablation}.
We can observe the overlapped jaggies in both the static regions (pointed by the yellow arrow) and the dynamic regions (pointed by the blue arrow) in temporal profiles of the ``baseline'' method, which indicates the flickering artifacts.
The ``+ initialization'' method can alleviate the flickering artifacts in the static regions, but still has jaggies in the dynamic regions, while our full method can produce smooth depth sequences in both static and dynamic regions.

\subsection{Applications}
\label{sec:applications}

Our DepthCrafter can facilitate various downstream applications, \eg, foreground matting, depth slicing, fog effects, and depth-conditioned video generation, by providing temporally consistent depth sequences with fine-grained details for open-world videos. 
We show example results of fog effects and depth-conditioned video generation in Fig.~\ref{fig:application}, while more visual effects results are available in the supplementary material.
For the fog effect, we blend the fog map with the input video frames based on the depth values to simulate varying transparency levels in fog.
And many recent conditioned video generation models~\cite{chen2023controlavideo, guo2023sparsectrl, feng2023ccedit, zhang2023controlvideo} employ depth maps as the structure conditions for video generation or editing.
We adopt Control-A-Video~\cite{chen2023controlavideo} and video depth of our method as conditions to generate a video with prompts ``\textit{a rider walking through stars, artstation}''.
The visual effects of these applications rely heavily on the accuracy and consistency of the video depth, which demonstrates the wide applicability of our DepthCrafter in various downstream tasks.

\begin{figure}
    \centering
    \includegraphics[width=1.0\linewidth]{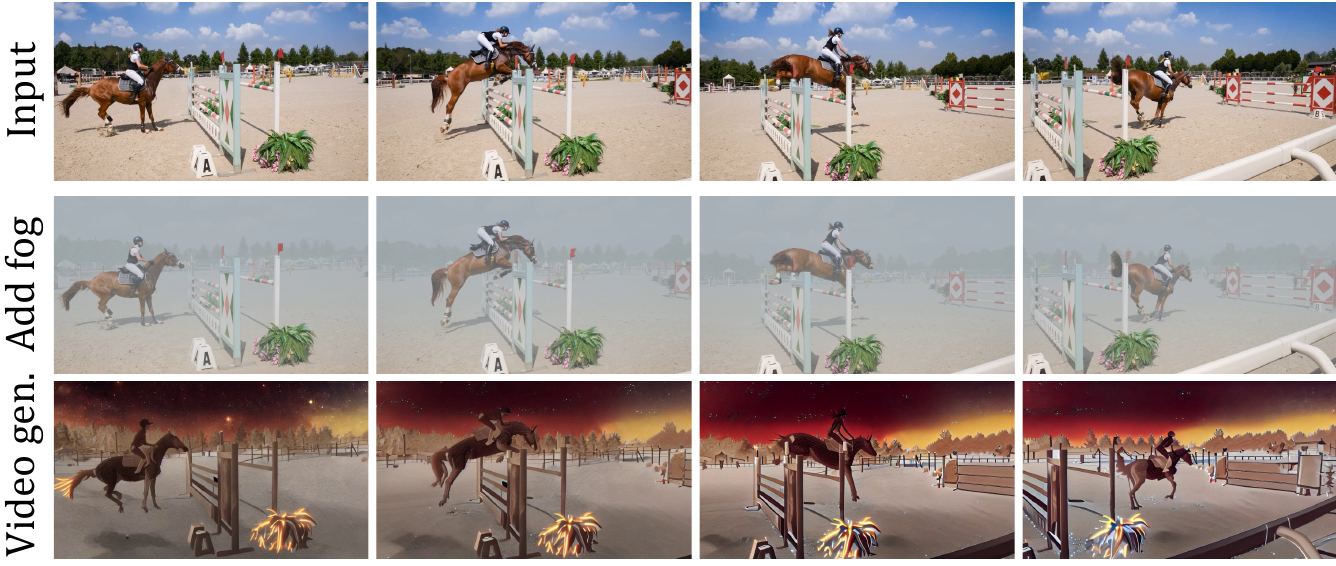}
    \vspace{-1.5em}
    \caption{Examples of visual effectiveness that could benefit from our DepthCrafter, including adding fog effects and depth-conditioned video generation. More visual effects results are available on our website.}
    \vspace{-0.5em}
    \label{fig:application}
\end{figure}

\subsection{Limitations}
\label{sec:limitations}
Although our DepthCrafter achieves state-of-the-art performance in open-world video depth estimation, there are still some limitations to be addressed in the future.
First, the computation and memory consumption of our model is relatively high, which is mainly due to the large model size and the iterative denoising process in the diffusion model.
As shown in Tab.~\ref{tab:speed}, we report the inference speed of our model, Depth-Anything (V2), and Marigold, with the resolution of $1024 \times 576$ on a single NVIDIA A100 GPU.
We can see that Depth-Anything (V2) is faster than our model, while our method is much faster than Marigold since it ensembles five inference results.
Our model achieves the best quality without ensembling, and its inference speed of 465.84 ms per frame is acceptable for many applications.
Second, our method requires around 24GB GPU memory to process a video with the resolution of $1024 \times 576$ and a segment length of 110 frames.
But note that, with the segment length of 40 frames, the memory consumption can be reduced to 12GB, which is acceptable for most modern GPUs, and we can still infer videos of any length by our inference strategy under this setting.
We believe that further engineering efforts, such as model distillation and quantization, can help further promote the practicality of our method.

\begin{table}[t]
    \small
        \centering
        \caption{Inference time per frame (ms) of our model, Depth-Anything (V2), and Marigold, with the resolution of $1024 \times 576$.}
        \vspace{-0.5em}
        \setlength\tabcolsep{1.5mm}
        \resizebox{\linewidth}{!}{
        \begin{tabular}{lcccc}
            \toprule
            Method & Encoding & Denoising & Decoding & All \\
            \midrule
            Depth-Anything (V2) & N/A & N/A & N/A & 180.46 \\
            Marigold & 256.40 & 114.53 & 699.36 & 1070.29 \\
            DepthCrafter (Ours) & 51.85 & 160.93 & 253.06 & 465.84 \\
            \bottomrule
        \end{tabular}
        }
        \vspace{-1.7em}
        \label{tab:speed}
\end{table}

\section{Conclusion} 
\label{sec:conclusion}
We present DepthCrafter, a novel open-world video depth estimation method that leverages video diffusion models.
It can generate temporally consistent depth sequences with fine-grained details for video width diverse content, motion, and camera movement, without requiring any additional information.
It also supports videos of variable lengths, ranging from one frame (static image) to extremely long videos.
This is achieved through our meticulously designed three-stage training strategy, compiled paired video-depth datasets, and an inference strategy.
Extensive evaluations have demonstrated that DepthCrafter achieves state-of-the-art performance in open-world video depth estimation under zero-shot settings.
It also facilitates various downstream applications, including depth-based visual effects and conditional video generation.

\clearpage

{
    \small
    \bibliographystyle{ieeenat_fullname}
    \bibliography{main}

\begin{thebibliography}{77}
\providecommand{\natexlab}[1]{#1}
\providecommand{\url}[1]{\texttt{#1}}
\expandafter\ifx\csname urlstyle\endcsname\relax
  \providecommand{\doi}[1]{doi: #1}\else
  \providecommand{\doi}{doi: \begingroup \urlstyle{rm}\Url}\fi

\bibitem[Aich et~al.(2021)Aich, Vianney, Islam, and Liu]{aich2021bidirectional}
Shubhra Aich, Jean Marie~Uwabeza Vianney, Md~Amirul Islam, and Mannat Kaur~Bingbing Liu.
\newblock Bidirectional attention network for monocular depth estimation.
\newblock In \emph{ICRA}, 2021.

\bibitem[Bhat et~al.(2023)Bhat, Birkl, Wofk, Wonka, and M{\"u}ller]{bhat2023zoedepth}
Shariq~Farooq Bhat, Reiner Birkl, Diana Wofk, Peter Wonka, and Matthias M{\"u}ller.
\newblock Zoedepth: Zero-shot transfer by combining relative and metric depth.
\newblock \emph{arXiv preprint arXiv:2302.12288}, 2023.

\bibitem[Blattmann et~al.(2023{\natexlab{a}})Blattmann, Dockhorn, Kulal, Mendelevitch, Kilian, Lorenz, Levi, English, Voleti, Letts, et~al.]{blattmann2023stable}
Andreas Blattmann, Tim Dockhorn, Sumith Kulal, Daniel Mendelevitch, Maciej Kilian, Dominik Lorenz, Yam Levi, Zion English, Vikram Voleti, Adam Letts, et~al.
\newblock Stable video diffusion: Scaling latent video diffusion models to large datasets.
\newblock \emph{arXiv preprint arXiv:2311.15127}, 2023{\natexlab{a}}.

\bibitem[Blattmann et~al.(2023{\natexlab{b}})Blattmann, Rombach, Ling, Dockhorn, Kim, Fidler, and Kreis]{blattmann2023align}
Andreas Blattmann, Robin Rombach, Huan Ling, Tim Dockhorn, Seung~Wook Kim, Sanja Fidler, and Karsten Kreis.
\newblock Align your latents: High-resolution video synthesis with latent diffusion models.
\newblock In \emph{CVPR}, 2023{\natexlab{b}}.

\bibitem[Bochkovskii et~al.(2024)Bochkovskii, Delaunoy, Germain, Santos, Zhou, Richter, and Koltun]{bochkovskii2024depth}
Aleksei Bochkovskii, Ama{\"e}l Delaunoy, Hugo Germain, Marcel Santos, Yichao Zhou, Stephan~R Richter, and Vladlen Koltun.
\newblock Depth pro: Sharp monocular metric depth in less than a second.
\newblock \emph{arXiv preprint arXiv:2410.02073}, 2024.

\bibitem[Brooks et~al.(2024)Brooks, Peebles, Holmes, DePue, Guo, Jing, Schnurr, Taylor, Luhman, Luhman, Ng, Wang, and Ramesh]{sora}
Tim Brooks, Bill Peebles, Connor Holmes, Will DePue, Yufei Guo, Li Jing, David Schnurr, Joe Taylor, Troy Luhman, Eric Luhman, Clarence Ng, Ricky Wang, and Aditya Ramesh.
\newblock Video generation models as world simulators, 2024.

\bibitem[Butler et~al.(2012)Butler, Wulff, Stanley, and Black]{Sintel}
D.~J. Butler, J. Wulff, G.~B. Stanley, and M.~J. Black.
\newblock A naturalistic open source movie for optical flow evaluation.
\newblock In \emph{ECCV}, 2012.

\bibitem[Chen et~al.(2023{\natexlab{a}})Chen, Xia, He, Zhang, Cun, Yang, Xing, Liu, Chen, Wang, et~al.]{chen2023videocrafter1}
Haoxin Chen, Menghan Xia, Yingqing He, Yong Zhang, Xiaodong Cun, Shaoshu Yang, Jinbo Xing, Yaofang Liu, Qifeng Chen, Xintao Wang, et~al.
\newblock Videocrafter1: Open diffusion models for high-quality video generation.
\newblock \emph{arXiv preprint arXiv:2310.19512}, 2023{\natexlab{a}}.

\bibitem[Chen et~al.(2024)Chen, Zhang, Cun, Xia, Wang, Weng, and Shan]{chen2024videocrafter2}
Haoxin Chen, Yong Zhang, Xiaodong Cun, Menghan Xia, Xintao Wang, Chao Weng, and Ying Shan.
\newblock Videocrafter2: Overcoming data limitations for high-quality video diffusion models.
\newblock In \emph{CVPR}, 2024.

\bibitem[Chen et~al.(2023{\natexlab{b}})Chen, Wu, Xie, Wu, Li, Xia, Xiao, and Lin]{chen2023controlavideo}
Weifeng Chen, Jie Wu, Pan Xie, Hefeng Wu, Jiashi Li, Xin Xia, Xuefeng Xiao, and Liang Lin.
\newblock Control-a-video: Controllable text-to-video generation with diffusion models, 2023{\natexlab{b}}.

\bibitem[Chen et~al.(2019)Chen, Schmid, and Sminchisescu]{chen2019self}
Yuhua Chen, Cordelia Schmid, and Cristian Sminchisescu.
\newblock Self-supervised learning with geometric constraints in monocular video: Connecting flow, depth, and camera.
\newblock In \emph{ICCV}, 2019.

\bibitem[Dai et~al.(2017)Dai, Chang, Savva, Halber, Funkhouser, and Nie{\ss}ner]{dai2017scannet}
Angela Dai, Angel~X. Chang, Manolis Savva, Maciej Halber, Thomas Funkhouser, and Matthias Nie{\ss}ner.
\newblock Scannet: Richly-annotated 3d reconstructions of indoor scenes.
\newblock In \emph{CVPR}, 2017.

\bibitem[Darcet et~al.(2024)Darcet, Oquab, Mairal, and Bojanowski]{darcet2023vitneedreg}
Timothée Darcet, Maxime Oquab, Julien Mairal, and Piotr Bojanowski.
\newblock Vision transformers need registers.
\newblock In \emph{ICLR}, 2024.

\bibitem[Dong et~al.(2022)Dong, Garratt, Anavatti, and Abbass]{dong2022towards}
Xingshuai Dong, Matthew~A Garratt, Sreenatha~G Anavatti, and Hussein~A Abbass.
\newblock Towards real-time monocular depth estimation for robotics: A survey.
\newblock \emph{IEEE Transactions on Intelligent Transportation Systems}, 23\penalty0 (10):\penalty0 16940--16961, 2022.

\bibitem[Eigen et~al.(2014)Eigen, Puhrsch, and Fergus]{eigen2014depth}
David Eigen, Christian Puhrsch, and Rob Fergus.
\newblock Depth map prediction from a single image using a multi-scale deep network.
\newblock \emph{NeurIPS}, 27, 2014.

\bibitem[Feng et~al.(2024)Feng, Weng, Wang, Yuan, Bao, Luo, Chen, and Guo]{feng2023ccedit}
Ruoyu Feng, Wenming Weng, Yanhui Wang, Yuhui Yuan, Jianmin Bao, Chong Luo, Zhibo Chen, and Baining Guo.
\newblock Ccedit: Creative and controllable video editing via diffusion models.
\newblock In \emph{CVPR}, 2024.

\bibitem[Fu et~al.(2018)Fu, Gong, Wang, Batmanghelich, and Tao]{fu2018deep}
Huan Fu, Mingming Gong, Chaohui Wang, Kayhan Batmanghelich, and Dacheng Tao.
\newblock Deep ordinal regression network for monocular depth estimation.
\newblock In \emph{CVPR}, 2018.

\bibitem[Fu et~al.(2024)Fu, Yin, Hu, Wang, Ma, Tan, Shen, Lin, and Long]{fu2024geowizard}
Xiao Fu, Wei Yin, Mu Hu, Kaixuan Wang, Yuexin Ma, Ping Tan, Shaojie Shen, Dahua Lin, and Xiaoxiao Long.
\newblock Geowizard: Unleashing the diffusion priors for 3d geometry estimation from a single image.
\newblock In \emph{ECCV}, 2024.

\bibitem[Geiger et~al.(2013)Geiger, Lenz, Stiller, and Urtasun]{Geiger2013IJRR}
Andreas Geiger, Philip Lenz, Christoph Stiller, and Raquel Urtasun.
\newblock Vision meets robotics: The kitti dataset.
\newblock \emph{IJRR}, 2013.

\bibitem[Guo et~al.(2024)Guo, Yang, Rao, Agrawala, Lin, and Dai]{guo2023sparsectrl}
Yuwei Guo, Ceyuan Yang, Anyi Rao, Maneesh Agrawala, Dahua Lin, and Bo Dai.
\newblock Sparsectrl: Adding sparse controls to text-to-video diffusion models.
\newblock In \emph{ECCV}, 2024.

\bibitem[He et~al.(2024)He, Li, Yin, Liang, Li, Zhou, Liu, Liu, and Chen]{he2024lotus}
Jing He, Haodong Li, Wei Yin, Yixun Liang, Leheng Li, Kaiqiang Zhou, Hongbo Liu, Bingbing Liu, and Ying-Cong Chen.
\newblock Lotus: Diffusion-based visual foundation model for high-quality dense prediction.
\newblock \emph{arXiv preprint arXiv:2409.18124}, 2024.

\bibitem[Ho et~al.(2020)Ho, Jain, and Abbeel]{ho2020denoising}
Jonathan Ho, Ajay Jain, and Pieter Abbeel.
\newblock Denoising diffusion probabilistic models.
\newblock \emph{NeurIPS}, 33:\penalty0 6840--6851, 2020.

\bibitem[Ho et~al.(2022)Ho, Salimans, Gritsenko, Chan, Norouzi, and Fleet]{ho2022video}
Jonathan Ho, Tim Salimans, Alexey Gritsenko, William Chan, Mohammad Norouzi, and David~J Fleet.
\newblock Video diffusion models.
\newblock \emph{NeurIPS}, 35:\penalty0 8633--8646, 2022.

\bibitem[Hong et~al.(2022)Hong, Zhang, Shen, and Xu]{hong2022depth}
Fa-Ting Hong, Longhao Zhang, Li Shen, and Dan Xu.
\newblock Depth-aware generative adversarial network for talking head video generation.
\newblock In \emph{CVPR}, 2022.

\bibitem[Hu et~al.(2024)Hu, Yin, Zhang, Cai, Long, Chen, Wang, Yu, Shen, and Shen]{hu2024metric3d}
Mu Hu, Wei Yin, Chi Zhang, Zhipeng Cai, Xiaoxiao Long, Hao Chen, Kaixuan Wang, Gang Yu, Chunhua Shen, and Shaojie Shen.
\newblock Metric3d v2: A versatile monocular geometric foundation model for zero-shot metric depth and surface normal estimation.
\newblock \emph{IEEE TPAMI}, 46\penalty0 (12):\penalty0 10579--10596, 2024.

\bibitem[Hu et~al.(2020)Hu, Xia, Fu, and Wong]{hu-2020-mononizing}
Wenbo Hu, Menghan Xia, Chi-Wing Fu, and Tien-Tsin Wong.
\newblock Mononizing binocular videos.
\newblock \emph{TOG (Proceedings of ACM SIGGRAPH Asia)}, 39\penalty0 (6):\penalty0 228:1--228:16, 2020.

\bibitem[Hu et~al.(2021)Hu, Zhao, Jiang, Jia, and Wong]{hu-2021-bidirectional}
Wenbo Hu, Hengshuang Zhao, Li Jiang, Jiaya Jia, and Tien-Tsin Wong.
\newblock Bidirectional projection network for cross dimensional scene understanding.
\newblock In \emph{CVPR}, 2021.

\bibitem[Hu et~al.(2023)Hu, Wang, Ma, Yang, Gao, Liu, and Ma]{hu2023Tri-MipRF}
Wenbo Hu, Yuling Wang, Lin Ma, Bangbang Yang, Lin Gao, Xiao Liu, and Yuewen Ma.
\newblock Tri-miprf: Tri-mip representation for efficient anti-aliasing neural radiance fields.
\newblock In \emph{ICCV}, 2023.

\bibitem[Jing et~al.(2024)Jing, Mao, and Mikolajczyk]{jing2024matchstereovideos}
Junpeng Jing, Ye Mao, and Krystian Mikolajczyk.
\newblock Match-stereo-videos: Bidirectional alignment for consistent dynamic stereo matching.
\newblock In \emph{ECCV}, 2024.

\bibitem[Karaev et~al.(2023)Karaev, Rocco, Graham, Neverova, Vedaldi, and Rupprecht]{karaev2023dynamicstereo}
Nikita Karaev, Ignacio Rocco, Benjamin Graham, Natalia Neverova, Andrea Vedaldi, and Christian Rupprecht.
\newblock Dynamicstereo: Consistent dynamic depth from stereo videos.
\newblock In \emph{CVPR}, 2023.

\bibitem[Karras et~al.(2022)Karras, Aittala, Aila, and Laine]{karras2022elucidating}
Tero Karras, Miika Aittala, Timo Aila, and Samuli Laine.
\newblock Elucidating the design space of diffusion-based generative models.
\newblock \emph{NeurIPS}, 35:\penalty0 26565--26577, 2022.

\bibitem[Ke et~al.(2024)Ke, Obukhov, Huang, Metzger, Daudt, and Schindler]{marigold}
Bingxin Ke, Anton Obukhov, Shengyu Huang, Nando Metzger, Rodrigo~Caye Daudt, and Konrad Schindler.
\newblock Repurposing diffusion-based image generators for monocular depth estimation.
\newblock In \emph{CVPR}, 2024.

\bibitem[Kingma(2013)]{kingma2013auto}
DP Kingma.
\newblock Auto-encoding variational bayes.
\newblock \emph{arXiv preprint arXiv:1312.6114}, 2013.

\bibitem[Kingma(2014)]{kingma2014adam}
Diederik~P Kingma.
\newblock Adam: A method for stochastic optimization.
\newblock \emph{arXiv preprint arXiv:1412.6980}, 2014.

\bibitem[Kopf et~al.(2021)Kopf, Rong, and Huang]{kopf2021rcvd}
Johannes Kopf, Xuejian Rong, and Jia-Bin Huang.
\newblock Robust consistent video depth estimation.
\newblock In \emph{CVPR}, 2021.

\bibitem[Lee et~al.(2019)Lee, Han, Ko, and Suh]{lee2019big}
Jin~Han Lee, Myung-Kyu Han, Dong~Wook Ko, and Il~Hong Suh.
\newblock From big to small: Multi-scale local planar guidance for monocular depth estimation.
\newblock \emph{arXiv preprint arXiv:1907.10326}, 2019.

\bibitem[Li et~al.(2023{\natexlab{a}})Li, Jiang, Xu, Xiangli, Wang, Lin, and Dai]{li2023matrixcity}
Yixuan Li, Lihan Jiang, Linning Xu, Yuanbo Xiangli, Zhenzhi Wang, Dahua Lin, and Bo Dai.
\newblock Matrixcity: A large-scale city dataset for city-scale neural rendering and beyond.
\newblock In \emph{ICCV}, 2023{\natexlab{a}}.

\bibitem[Li et~al.(2023{\natexlab{b}})Li, Chen, Liu, and Jiang]{li2023depthformer}
Zhenyu Li, Zehui Chen, Xianming Liu, and Junjun Jiang.
\newblock Depthformer: Exploiting long-range correlation and local information for accurate monocular depth estimation.
\newblock \emph{Machine Intelligence Research}, 20\penalty0 (6):\penalty0 837--854, 2023{\natexlab{b}}.

\bibitem[Li et~al.(2023{\natexlab{c}})Li, Ye, Wang, Creighton, Taylor, Venkatesh, and Unberath]{li2023temporally}
Zhaoshuo Li, Wei Ye, Dilin Wang, Francis~X Creighton, Russell~H Taylor, Ganesh Venkatesh, and Mathias Unberath.
\newblock Temporally consistent online depth estimation in dynamic scenes.
\newblock In \emph{WACV}, 2023{\natexlab{c}}.

\bibitem[Liu et~al.(2021)Liu, Qi, and Fu]{liu20213d}
Zhengzhe Liu, Xiaojuan Qi, and Chi-Wing Fu.
\newblock 3d-to-2d distillation for indoor scene parsing.
\newblock In \emph{CVPR}, 2021.

\bibitem[Luo et~al.(2020)Luo, Huang, Szeliski, Matzen, and Kopf]{Luo-VideoDepth-2020}
Xuan Luo, Jia{-}Bin Huang, Richard Szeliski, Kevin Matzen, and Johannes Kopf.
\newblock Consistent video depth estimation.
\newblock \emph{TOG (Proceedings of ACM SIGGRAPH)}, 39\penalty0 (4), 2020.

\bibitem[Martin~Garcia et~al.(2025)Martin~Garcia, Abou~Zeid, Schmidt, de~Geus, Hermans, and Leibe]{garcia2024fine}
Gonzalo Martin~Garcia, Karim Abou~Zeid, Christian Schmidt, Daan de Geus, Alexander Hermans, and Bastian Leibe.
\newblock Fine-tuning image-conditional diffusion models is easier than you think.
\newblock In \emph{WACV}, 2025.

\bibitem[Oquab et~al.(2024)Oquab, Darcet, Moutakanni, Vo, Szafraniec, Khalidov, Fernandez, Haziza, Massa, El-Nouby, Howes, Huang, Xu, Sharma, Li, Galuba, Rabbat, Assran, Ballas, Synnaeve, Misra, Jegou, Mairal, Labatut, Joulin, and Bojanowski]{oquab2023dinov2}
Maxime Oquab, Timothée Darcet, Theo Moutakanni, Huy~V. Vo, Marc Szafraniec, Vasil Khalidov, Pierre Fernandez, Daniel Haziza, Francisco Massa, Alaaeldin El-Nouby, Russell Howes, Po-Yao Huang, Hu Xu, Vasu Sharma, Shang-Wen Li, Wojciech Galuba, Mike Rabbat, Mido Assran, Nicolas Ballas, Gabriel Synnaeve, Ishan Misra, Herve Jegou, Julien Mairal, Patrick Labatut, Armand Joulin, and Piotr Bojanowski.
\newblock Dinov2: Learning robust visual features without supervision.
\newblock In \emph{TMLR}, 2024.

\bibitem[Palazzolo et~al.(2019)Palazzolo, Behley, Lottes, Gigu\`ere, and Stachniss]{palazzolo2019iros}
E. Palazzolo, J. Behley, P. Lottes, P. Gigu\`ere, and C. Stachniss.
\newblock {ReFusion: 3D Reconstruction in Dynamic Environments for RGB-D Cameras Exploiting Residuals}.
\newblock In \emph{IROS}, 2019.

\bibitem[Patil et~al.(2022)Patil, Sakaridis, Liniger, and Van~Gool]{patil2022p3depth}
Vaishakh Patil, Christos Sakaridis, Alexander Liniger, and Luc Van~Gool.
\newblock P3depth: Monocular depth estimation with a piecewise planarity prior.
\newblock In \emph{CVPR}, 2022.

\bibitem[Perazzi et~al.(2016)Perazzi, Pont-Tuset, McWilliams, {Van Gool}, Gross, and Sorkine-Hornung]{Perazzi2016}
F. Perazzi, J. Pont-Tuset, B. McWilliams, L. {Van Gool}, M. Gross, and A. Sorkine-Hornung.
\newblock A benchmark dataset and evaluation methodology for video object segmentation.
\newblock In \emph{CVPR}, 2016.

\bibitem[Piccinelli et~al.(2024)Piccinelli, Yang, Sakaridis, Segu, Li, Van~Gool, and Yu]{piccinelli2024unidepth}
Luigi Piccinelli, Yung-Hsu Yang, Christos Sakaridis, Mattia Segu, Siyuan Li, Luc Van~Gool, and Fisher Yu.
\newblock Unidepth: Universal monocular metric depth estimation.
\newblock In \emph{CVPR}, 2024.

\bibitem[Radford et~al.(2021)Radford, Kim, Hallacy, Ramesh, Goh, Agarwal, Sastry, Askell, Mishkin, Clark, et~al.]{radford2021learning}
Alec Radford, Jong~Wook Kim, Chris Hallacy, Aditya Ramesh, Gabriel Goh, Sandhini Agarwal, Girish Sastry, Amanda Askell, Pamela Mishkin, Jack Clark, et~al.
\newblock Learning transferable visual models from natural language supervision.
\newblock In \emph{ICML}, 2021.

\bibitem[Ranftl et~al.(2020)Ranftl, Lasinger, Hafner, Schindler, and Koltun]{ranftl2020towards}
Ren{\'e} Ranftl, Katrin Lasinger, David Hafner, Konrad Schindler, and Vladlen Koltun.
\newblock Towards robust monocular depth estimation: Mixing datasets for zero-shot cross-dataset transfer.
\newblock \emph{IEEE TPAMI}, 44\penalty0 (3):\penalty0 1623--1637, 2020.

\bibitem[Ranftl et~al.(2021)Ranftl, Bochkovskiy, and Koltun]{ranftl2021vision}
Ren{\'e} Ranftl, Alexey Bochkovskiy, and Vladlen Koltun.
\newblock Vision transformers for dense prediction.
\newblock In \emph{ICCV}, 2021.

\bibitem[Rombach et~al.(2022)Rombach, Blattmann, Lorenz, Esser, and Ommer]{rombach2022high}
Robin Rombach, Andreas Blattmann, Dominik Lorenz, Patrick Esser, and Bj{\"o}rn Ommer.
\newblock High-resolution image synthesis with latent diffusion models.
\newblock In \emph{CVPR}, 2022.

\bibitem[Ronneberger et~al.(2015)Ronneberger, Fischer, and Brox]{ronneberger2015u}
Olaf Ronneberger, Philipp Fischer, and Thomas Brox.
\newblock U-net: Convolutional networks for biomedical image segmentation.
\newblock In \emph{MICCAI}, 2015.

\bibitem[Salimans and Ho(2022)]{salimans2022progressive}
Tim Salimans and Jonathan Ho.
\newblock Progressive distillation for fast sampling of diffusion models.
\newblock In \emph{ICLR}, 2022.

\bibitem[Shao et~al.(2024)Shao, Yang, Zhou, Zhang, Shen, Poggi, and Liao]{shao2024learning}
Jiahao Shao, Yuanbo Yang, Hongyu Zhou, Youmin Zhang, Yujun Shen, Matteo Poggi, and Yiyi Liao.
\newblock Learning temporally consistent video depth from video diffusion priors.
\newblock \emph{arXiv preprint arXiv:2406.01493}, 2024.

\bibitem[Silberman et~al.(2012)Silberman, Hoiem, Kohli, and Fergus]{silberman2012indoor}
Nathan Silberman, Derek Hoiem, Pushmeet Kohli, and Rob Fergus.
\newblock Indoor segmentation and support inference from rgbd images.
\newblock In \emph{ECCV}, 2012.

\bibitem[Sohl-Dickstein et~al.(2015)Sohl-Dickstein, Weiss, Maheswaranathan, and Ganguli]{sohl2015deep}
Jascha Sohl-Dickstein, Eric Weiss, Niru Maheswaranathan, and Surya Ganguli.
\newblock Deep unsupervised learning using nonequilibrium thermodynamics.
\newblock In \emph{ICML}, 2015.

\bibitem[Sun et~al.(2021)Sun, Xie, Chen, Zhou, and Bao]{sun2021neuralrecon}
Jiaming Sun, Yiming Xie, Linghao Chen, Xiaowei Zhou, and Hujun Bao.
\newblock Neuralrecon: Real-time coherent 3d reconstruction from monocular video.
\newblock In \emph{CVPR}, 2021.

\bibitem[Teed and Deng(2020)]{teed2018deepv2d}
Zachary Teed and Jia Deng.
\newblock Deepv2d: Video to depth with differentiable structure from motion.
\newblock In \emph{ICLR}, 2020.

\bibitem[von Platen et~al.(2022)von Platen, Patil, Lozhkov, Cuenca, Lambert, Rasul, Davaadorj, Nair, Paul, Berman, Xu, Liu, and Wolf]{von-platen-etal-2022-diffusers}
Patrick von Platen, Suraj Patil, Anton Lozhkov, Pedro Cuenca, Nathan Lambert, Kashif Rasul, Mishig Davaadorj, Dhruv Nair, Sayak Paul, William Berman, Yiyi Xu, Steven Liu, and Thomas Wolf.
\newblock Diffusers: State-of-the-art diffusion models.
\newblock \url{https://github.com/huggingface/diffusers}, 2022.

\bibitem[Wang et~al.(2019)Wang, Lucey, Perazzi, and Wang]{wang2019web}
Chaoyang Wang, Simon Lucey, Federico Perazzi, and Oliver Wang.
\newblock Web stereo video supervision for depth prediction from dynamic scenes.
\newblock In \emph{IEEE 3DV}, 2019.

\bibitem[Wang et~al.(2022)Wang, Pan, Li, Cao, Xian, and Zhang]{wang2022less}
Yiran Wang, Zhiyu Pan, Xingyi Li, Zhiguo Cao, Ke Xian, and Jianming Zhang.
\newblock Less is more: Consistent video depth estimation with masked frames modeling.
\newblock In \emph{ACM MM}, 2022.

\bibitem[Wang et~al.(2023{\natexlab{a}})Wang, Shi, Li, Huang, Cao, Zhang, Xian, and Lin]{Wang_2023_ICCV}
Yiran Wang, Min Shi, Jiaqi Li, Zihao Huang, Zhiguo Cao, Jianming Zhang, Ke Xian, and Guosheng Lin.
\newblock Neural video depth stabilizer.
\newblock In \emph{ICCV}, 2023{\natexlab{a}}.

\bibitem[Wang et~al.(2023{\natexlab{b}})Wang, Shi, Li, Huang, Cao, Zhang, Xian, and Lin]{wang2023neural}
Yiran Wang, Min Shi, Jiaqi Li, Zihao Huang, Zhiguo Cao, Jianming Zhang, Ke Xian, and Guosheng Lin.
\newblock Neural video depth stabilizer.
\newblock In \emph{ICCV}, 2023{\natexlab{b}}.

\bibitem[Xing et~al.(2024{\natexlab{a}})Xing, Liu, Xia, Zhang, Wang, Shan, and Wong]{xing2024tooncrafter}
Jinbo Xing, Hanyuan Liu, Menghan Xia, Yong Zhang, Xintao Wang, Ying Shan, and Tien-Tsin Wong.
\newblock Tooncrafter: Generative cartoon interpolation.
\newblock \emph{TOG (Proceedings of ACM SIGGRAPH Asia)}, 2024{\natexlab{a}}.

\bibitem[Xing et~al.(2024{\natexlab{b}})Xing, Xia, Zhang, Chen, Yu, Liu, Wang, Wong, and Shan]{xing2023dynamicrafter}
Jinbo Xing, Menghan Xia, Yong Zhang, Haoxin Chen, Wangbo Yu, Hanyuan Liu, Xintao Wang, Tien-Tsin Wong, and Ying Shan.
\newblock Dynamicrafter: Animating open-domain images with video diffusion priors.
\newblock In \emph{ECCV}, 2024{\natexlab{b}}.

\bibitem[Yang et~al.(2021)Yang, Tang, Ding, Sebe, and Ricci]{yang2021transformer}
Guanglei Yang, Hao Tang, Mingli Ding, Nicu Sebe, and Elisa Ricci.
\newblock Transformer-based attention networks for continuous pixel-wise prediction.
\newblock In \emph{ICCV}, 2021.

\bibitem[Yang et~al.(2024{\natexlab{a}})Yang, Kang, Huang, Xu, Feng, and Zhao]{depthanything}
Lihe Yang, Bingyi Kang, Zilong Huang, Xiaogang Xu, Jiashi Feng, and Hengshuang Zhao.
\newblock Depth anything: Unleashing the power of large-scale unlabeled data.
\newblock In \emph{CVPR}, 2024{\natexlab{a}}.

\bibitem[Yang et~al.(2024{\natexlab{b}})Yang, Kang, Huang, Zhao, Xu, Feng, and Zhao]{depth_anything_v2}
Lihe Yang, Bingyi Kang, Zilong Huang, Zhen Zhao, Xiaogang Xu, Jiashi Feng, and Hengshuang Zhao.
\newblock Depth anything v2.
\newblock \emph{NeurIPS}, 2024{\natexlab{b}}.

\bibitem[Yasarla et~al.(2023)Yasarla, Cai, Jeong, Shi, Garrepalli, and Porikli]{yasarla2023mamo}
Rajeev Yasarla, Hong Cai, Jisoo Jeong, Yunxiao Shi, Risheek Garrepalli, and Fatih Porikli.
\newblock Mamo: Leveraging memory and attention for monocular video depth estimation.
\newblock In \emph{ICCV}, 2023.

\bibitem[Yasarla et~al.(2024)Yasarla, Singh, Cai, Shi, Jeong, Zhu, Han, Garrepalli, and Porikli]{yasarla2024futuredepth}
Rajeev Yasarla, Manish~Kumar Singh, Hong Cai, Yunxiao Shi, Jisoo Jeong, Yinhao Zhu, Shizhong Han, Risheek Garrepalli, and Fatih Porikli.
\newblock Futuredepth: Learning to predict the future improves video depth estimation.
\newblock In \emph{ECCV}, 2024.

\bibitem[Yin et~al.(2023)Yin, Zhang, Chen, Cai, Yu, Wang, Chen, and Shen]{yin2023metric3d}
Wei Yin, Chi Zhang, Hao Chen, Zhipeng Cai, Gang Yu, Kaixuan Wang, Xiaozhi Chen, and Chunhua Shen.
\newblock Metric3d: Towards zero-shot metric 3d prediction from a single image.
\newblock In \emph{ICCV}, 2023.

\bibitem[Yu et~al.(2024)Yu, Xing, Yuan, Hu, Li, Huang, Gao, Wong, Shan, and Tian]{yu2024viewcrafter}
Wangbo Yu, Jinbo Xing, Li Yuan, Wenbo Hu, Xiaoyu Li, Zhipeng Huang, Xiangjun Gao, Tien-Tsin Wong, Ying Shan, and Yonghong Tian.
\newblock Viewcrafter: Taming video diffusion models for high-fidelity novel view synthesis.
\newblock \emph{arXiv preprint arXiv:2409.02048}, 2024.

\bibitem[Yu et~al.(2022)Yu, Peng, Niemeyer, Sattler, and Geiger]{Yu2022MonoSDF}
Zehao Yu, Songyou Peng, Michael Niemeyer, Torsten Sattler, and Andreas Geiger.
\newblock Monosdf: Exploring monocular geometric cues for neural implicit surface reconstruction.
\newblock \emph{NeurIPS}, 2022.

\bibitem[Zhang et~al.(2019)Zhang, Shen, Li, Cao, Liu, and Yan]{zhang2019exploiting}
Haokui Zhang, Chunhua Shen, Ying Li, Yuanzhouhan Cao, Yu Liu, and Youliang Yan.
\newblock Exploiting temporal consistency for real-time video depth estimation.
\newblock In \emph{ICCV}, 2019.

\bibitem[Zhang et~al.(2024{\natexlab{a}})Zhang, Gu, Wang, Wang, Cheng, Zhu, and Zou]{zhang2024mimicmotion}
Yuang Zhang, Jiaxi Gu, Li-Wen Wang, Han Wang, Junqi Cheng, Yuefeng Zhu, and Fangyuan Zou.
\newblock Mimicmotion: High-quality human motion video generation with confidence-aware pose guidance.
\newblock \emph{arXiv preprint arXiv:2406.19680}, 2024{\natexlab{a}}.

\bibitem[Zhang et~al.(2024{\natexlab{b}})Zhang, Wei, Jiang, Zhang, Zuo, and Tian]{zhang2023controlvideo}
Yabo Zhang, Yuxiang Wei, Dongsheng Jiang, Xiaopeng Zhang, Wangmeng Zuo, and Qi Tian.
\newblock Controlvideo: Training-free controllable text-to-video generation.
\newblock In \emph{ICLR}, 2024{\natexlab{b}}.

\bibitem[Zhang et~al.(2021)Zhang, Cole, Tucker, Freeman, and Dekel]{zhang2021consistent}
Zhoutong Zhang, Forrester Cole, Richard Tucker, William~T Freeman, and Tali Dekel.
\newblock Consistent depth of moving objects in video.
\newblock \emph{TOG (Proceedings of ACM SIGGRAPH)}, 40\penalty0 (4):\penalty0 1--12, 2021.

\end{thebibliography}
}

\clearpage

\appendix

\renewcommand*{\thesection}{\Alph{section}}
\newcommand{\multiref}[2]{\cref{#1}--\ref{#2}}
\renewcommand{\thetable}{S\arabic{table}}
\renewcommand{\thefigure}{S\arabic{figure}}

\setcounter{table}{0}
\setcounter{figure}{0}
\setcounter{page}{1}

\maketitlesupplementary

\section*{\Large Appendix}

\noindent
In this supplementary material, we provide additional implementation details in~\cref{suppsec:implementation} and more evaluations in~\cref{suppsec:quantitative}.
For more visual results and details, \emph{we highly recommend referring to the webpage: \url{https://depthcrafter.github.io}}, since the visual quality of the generated depth sequences can be better accessed with interactive videos.
We would release the code and model to facilitate further research and applications.

\section{Implementation Details}
\label{suppsec:implementation}

\subsection{Data Preparation}
\label{suppsec:data}
Following conventional practice in depth estimation~\cite{depthanything,depth_anything_v2}, we represent the depth in the disparity domain.
We target relative depth estimation, so we normalize the disparity values to the range of $[0, 1]$ by the maximum and minimum disparity values in the sequence.
Since the training of our DepthCrafter only involves the U-Net model, with the VAE freezing, we can pre-process the latents of videos and the corresponding depth sequences in advance.
This caching mechanism significantly reduces the training time and memory consumption, as the latents do not need to be re-computed during the training process, and the VAE does not need to be loaded into the memory.

\subsection{Training Details}
\label{suppsec:training}
We follow the EDM-framework~\cite{karras2022elucidating} to train our DepthCrafter.
This can be mathematically formulated as~\cref{eq:diffusion,eq:denoising_loss,eq:denoiser}.
The $c_\text{in}$, $c_\text{out}$, $c_\text{skip}$, and $c_\text{noise}$ in~\cref{eq:denoiser} are the EDM preconditionining functions~\cite{karras2022elucidating,salimans2022progressive}:
\begin{equation}
    \begin{split}
        c_\text{in} (\sigma_t) &= 1 / \sqrt{1 + \sigma_t^2}, \\
        c_\text{out} (\sigma_t) &= -\sigma_t / \sqrt{1 + \sigma_t^2}, \\
        c_\text{skip} (\sigma_t) &= 1 / (1 + \sigma_t^2), \\
        c_\text{noise} (\sigma_t) &= 0.25 \cdot \log(\sigma_t).
    \end{split} 
\end{equation}
$c_\text{in}$ and $c_\text{out}$ are used to scale the input and output magnitudes, $c_\text{skip}$ is used to modulate the skip connection, and $c_\text{noise}$ is used to map the noise level $\sigma_t$ into a conditioning input for the denoiser $F_\theta$. 
The $\lambda_{\sigma_t}$ in~\cref{eq:denoising_loss} effectively incurs a per-sample loss weight for balancing different noise levels, which is set as:
\begin{equation}
    \lambda_{\sigma_t} = 1 / c_\text{out}(\sigma_t)^2.
\end{equation}
During training, we randomly sample the noise level $\sigma_t$ from a log-normal distribution:
\begin{equation}
    \text{ln}(\sigma_t) \sim \mathcal{N}(0.7, 1.6^2),
\end{equation} 
which is following the EDM-framework~\cite{karras2022elucidating} to target the training efforts to the relevant range.

We train our DepthCrafter on eight NVIDIA A100 GPUs with a learning rate of $10^{-5}$, and a batch size of 8.
We adopted the DeepSpeed ZeRO-2 strategy, gradient checkpointing, and mixed precision training to reduce memory consumption during training.
We also highly optimize the U-Net structure and cache the latents to further reduce memory consumption.
The first and third training stages consume around 40GB of GPU memory per device, while the second stage consumes around 80GB.
The ``temporal layers'' mentioned in~\cref{sec:method} are the layers performed on the time axis, such as temporal transformer and temporal resnet blocks.
The remaining layers are spatial layers, such as spatial transformers and spatial resnet blocks.

\begin{figure*}[!t]
    \centering
    \includegraphics[width=1.0\linewidth]{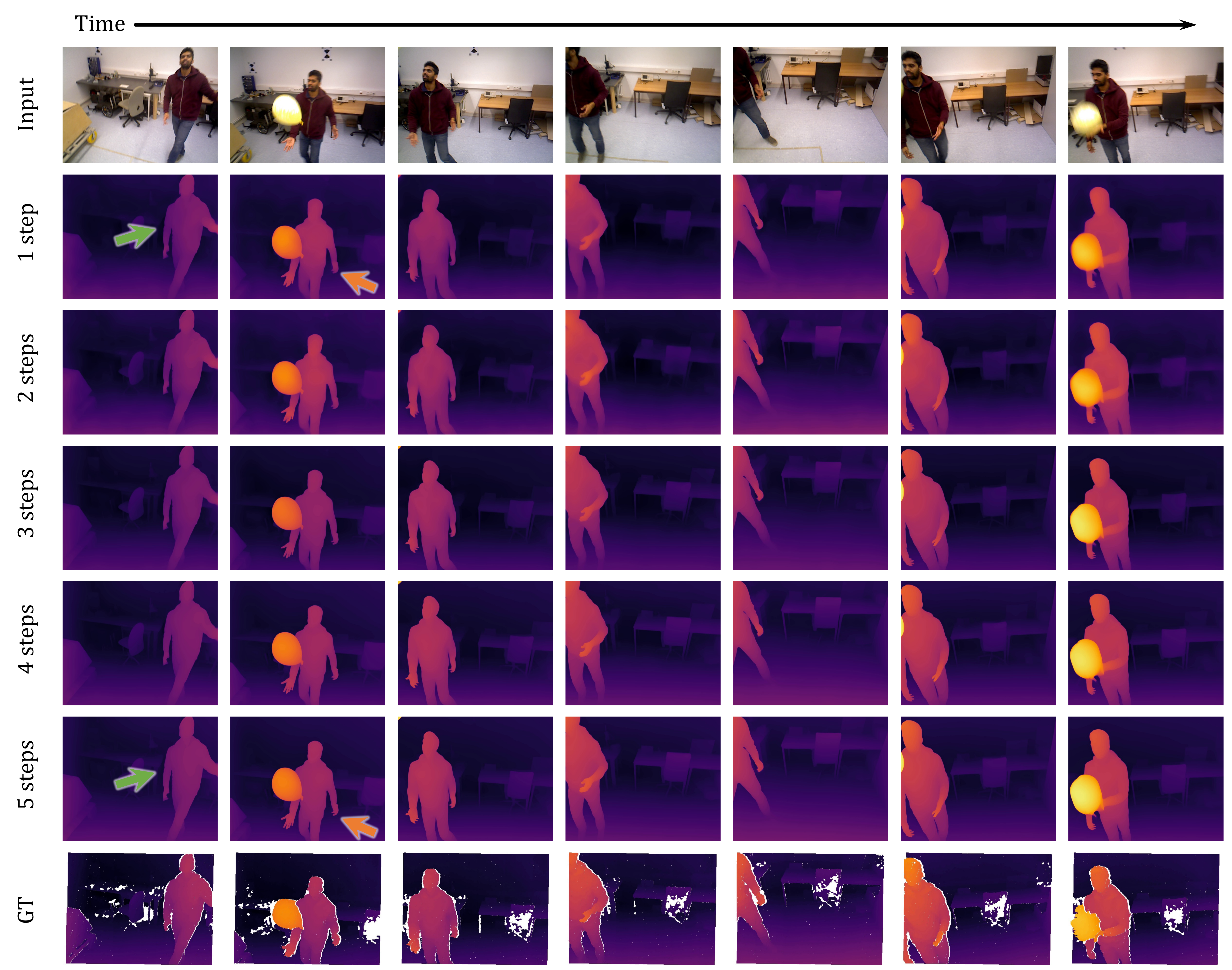}
    \caption{
   Effects of the number of denoising steps in our DepthCrafter. We show an example from the Bonn dataset~\cite{palazzolo2019iros}, where the depth sequences are generated with different numbers of denoising steps. The green and orange arrows indicate the regions where more denoising steps can refine the depth details.
    }
    \label{suppfig:steps}
\end{figure*}

\begin{table*}[!t]
    \centering
    \caption{
        Performance comparison of our DepthCrafter with different numbers of denoising steps. For reference, we also include the results of Marigold~\cite{marigold} and Depth-Anything-V2~\cite{depth_anything_v2}. The inference speed is measured in milliseconds per frame at the resolution of 1024$\times$576. \textbf{Best} and \underline{second best} results are highlighted.
        }
    \resizebox{\textwidth}{!}{
        \begin{tabular}{l|c|c|cc|cc|cc|cc}
            \toprule
            \multirow{2}{*}{\textbf{Method}} & \multirow{2}{*}{\textbf{Steps}} & \textbf{\textit{ms} / frame $\downarrow$} & \multicolumn{2}{c}{\textbf{Sintel} ($\sim$50 frames)} & \multicolumn{2}{c}{\textbf{Scannet} (90 frames)} & \multicolumn{2}{c}{\textbf{KITTI} (110 frames)} & \multicolumn{2}{c}{\textbf{Bonn} (110 frames)} \\
            \cmidrule(lr){4-5}\cmidrule(lr){6-7}\cmidrule(lr){8-9}\cmidrule(lr){10-11}
            & & @1024$\times$576 & \textbf{AbsRel}$\downarrow$ & $\boldsymbol{\delta}_1$$\uparrow$ & \textbf{AbsRel}$\downarrow$ & $\boldsymbol{\delta}_1$$\uparrow$ & \textbf{AbsRel}$\downarrow$ & $\boldsymbol{\delta}_1$$\uparrow$ & \textbf{AbsRel}$\downarrow$ & $\boldsymbol{\delta}_1$$\uparrow$ \\
            \midrule
            Marigold~\cite{marigold} &  & 1070.29 & 0.532 & 0.515 &  0.166 & 0.769 & 0.149 & 0.796 & 0.091 & 0.931  \\
            Depth-Anything-V2~\cite{depth_anything_v2} &  & \textbf{180.46} & 0.367 & 0.554 & 0.135 & 0.822 & 0.140 & 0.804 & 0.106 & 0.921 \\
            \midrule
            \multirow{8}{*}{DepthCrafter (Ours)} 
            & 1 & \underline{337.10} & 0.319 & 0.651 & 0.132 & 0.826 & 0.138 & 0.812 & 0.084 & 0.954 \\   
            & 2 & 369.28 & 0.301 & 0.661 & 0.132 & 0.828 & 0.138 & 0.814 & 0.083 & 0.955 \\   
            & 3 & 401.47 & \underline{0.273} & 0.693 & \textbf{0.123} & \underline{0.854} & 0.111 & 0.877 & 0.073 & 0.971 \\   
            & 4 & 433.65 & 0.293 & \textbf{0.697} & \textbf{0.123} & \textbf{0.856} & 0.107 & 0.888 & \underline{0.072} & 0.971 \\   
            & 5 & 465.84 & \textbf{0.270} & \textbf{0.697} & \textbf{0.123} & \textbf{0.856} & \textbf{0.104} & \textbf{0.896} & \textbf{0.071} & \underline{0.972} \\   
            & 6 & 498.03 & 0.299 & \underline{0.696} & \underline{0.124} & 0.851 & \underline{0.105} & \underline{0.891} & \underline{0.072} & \textbf{0.973} \\   
            & 10 & 626.72 & 0.291 & 0.694 & 0.125 & 0.849 & 0.106 & 0.890 & 0.073 & \underline{0.972} \\   
            & 25 & 1109.42 & 0.292 & \textbf{0.697} & 0.125 & 0.848 & 0.110 & 0.881 & 0.075 & 0.971 \\    
            \bottomrule
        \end{tabular}
    }
    \label{supptab:zeroshot_mde}
\end{table*}

\begin{table*}[!t]
    \centering
    \caption{
        Effectiveness of our three-stage training strategy. We show the performance of our DepthCrafter with different training stages on the Sintel, Scannet, KITTI, and Bonn datasets. For reference, we also include the results of Marigold~\cite{marigold} and Depth-Anything-V2~\cite{depth_anything_v2}. \textbf{Best} and \underline{second best} results are highlighted.
        }
    \resizebox{\textwidth}{!}{
        \begin{tabular}{l|c|cc|cc|cc|cc}
        \toprule
        \multirow{2}{*}{\textbf{Method}} & \multirow{2}{*}{\textbf{Training Stages}} & \multicolumn{2}{c}{\textbf{Sintel} ($\sim$50 frames)} & \multicolumn{2}{c}{\textbf{Scannet} (90 frames)} & \multicolumn{2}{c}{\textbf{KITTI} (110 frames)} & \multicolumn{2}{c}{\textbf{Bonn} (110 frames)} \\
        \cmidrule(lr){3-4}\cmidrule(lr){5-6}\cmidrule(lr){7-8}\cmidrule(lr){9-10}
        & & \textbf{AbsRel}$\downarrow$ & $\boldsymbol{\delta}_1$$\uparrow$ &  \textbf{AbsRel}$\downarrow$ & $\boldsymbol{\delta}_1$$\uparrow$ &  \textbf{AbsRel}$\downarrow$ & $\boldsymbol{\delta}_1$$\uparrow$ &  \textbf{AbsRel}$\downarrow$ & $\boldsymbol{\delta}_1$$\uparrow$ \\
        \midrule
        Marigold~\cite{marigold} &  & 0.532 & 0.515 & 0.166 & 0.769 & 0.149 & 0.796 & 0.091 & 0.931  \\
        Depth-Anything-V2~\cite{depth_anything_v2} &  & 0.367 & 0.554 & 0.135 & 0.822 & 0.140 & 0.804 & 0.106 & 0.921 \\
        \midrule
        \multirow{3}{*}{DepthCrafter (Ours)} 
        & 1 & 0.322 & 0.626 & 0.170 & 0.721 & 0.174 & 0.724 & 0.103 & 0.917 \\   
        & 2 & \underline{0.316} & \underline{0.675} & \underline{0.134} & \underline{0.826} & \underline{0.127} & \underline{0.844} & \underline{0.090} & \underline{0.935} \\   
        & 3 & \textbf{0.270} & \textbf{0.697} & \textbf{0.123} & \textbf{0.856} & \textbf{0.104} & \textbf{0.896} & \textbf{0.071} & \textbf{0.972} \\   
        \bottomrule
        \end{tabular}
    }
    \label{supptab:training_stages}
\end{table*}

\subsection{Benchmark Evaluation Details}
\label{suppsec:benchmark}
Since existing monocular depth estimation methods and benchmarks are mainly tailored for static images, we re-compile the public benchmarks to evaluate the video depth estimation methods.
First, we re-format the testing datasets in the form of videos that are originally in the form of images.
Specifically, for the ScanNet V2 dataset~\cite{dai2017scannet}, we extracted the first 90 RGB-D frames from the original sensor data sequences at a rate of 15 frames per second.
Besides, there are black regions in the corners of the images due to the camera calibration, which would affect the depth estimation evaluation.
We carefully crop the borders of the images to remove the black regions, \ie cropping 8 pixels from the top and bottom, and 11 pixels from the left and right.
For the KITTI dataset~\cite{Geiger2013IJRR}, we extracted the first 110 frames from the original sequences without downsampling the frame rate, since the difference between consecutive frames is relatively large.
And for the Bonn dataset~\cite{palazzolo2019iros}, which was usually not included in the evaluation due to the small scale, but we find it is a good complement to the ScanNet dataset for indoor scenes as it contains dynamic contents while ScanNet contains only static scenes.
We selected five sequences from the Bonn dataset, each with 110 frames, for evaluation.
For the Sintel dataset~\cite{Sintel}, as it is a synthetic dataset, we directly used the original sequences.

For the evaluation metrics, we followed the insight from the commonly used metrics in the depth estimation, including the absolute relative error (AbsRel) and the $\delta_1$ metric, but modified the scale and shift alignment from per-image to per-video.
This is because the depth values for a video should be consistent across frames, otherwise, the depth sequences would be flickering.
During evaluation, we first align the depth sequences to the ground truth by the scale and shift, using a least-square optimization.
Following MiDas~\cite{ranftl2020towards}, we cap the maximum depth values to a certain value for different datasets, \eg, 70 meters for the SinTel dataset, 80 meters for the KITTI dataset, and 10 meters for the ScanNet, Bonn, and NYUv2 datasets.

\begin{figure*}[!t]
    \centering
    \includegraphics[width=1.0\linewidth]{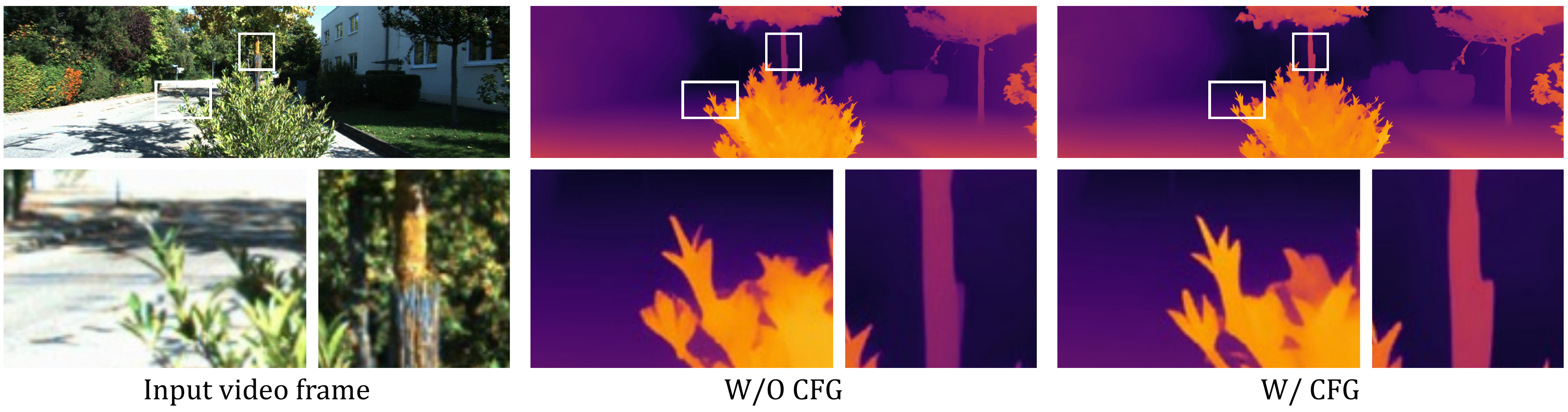}
    \caption{
    Effects of the classifier-free guidance (CFG) on the generated depth sequences. For better visualization, we blow up two regions that contain structures with fine-grained details. 
    }
    \label{suppfig:CFG_scale}
\end{figure*}

\begin{table*}[!t]
    \centering
    \caption{
        Effects of the classifier-free guidance (CFG). For reference, we also include the results of Marigold~\cite{marigold} and Depth-Anything-V2~\cite{depth_anything_v2}. \textbf{Best} and \underline{second best} results are highlighted.
        }
    \resizebox{\textwidth}{!}{
        \begin{tabular}{l|cc|cc|cc|cc}
            \toprule
            \multirow{2}{*}{\textbf{Method}} & \multicolumn{2}{c}{\textbf{Sintel} ($\sim$50 frames)} & \multicolumn{2}{c}{\textbf{Scannet} (90 frames)} & \multicolumn{2}{c}{\textbf{KITTI} (110 frames)} & \multicolumn{2}{c}{\textbf{Bonn} (110 frames)} \\
            \cmidrule(lr){2-3}\cmidrule(lr){4-5}\cmidrule(lr){6-7}\cmidrule(lr){8-9}
            & \textbf{AbsRel}$\downarrow$ & $\boldsymbol{\delta}_1$$\uparrow$ & \textbf{AbsRel}$\downarrow$ & $\boldsymbol{\delta}_1$$\uparrow$ & \textbf{AbsRel}$\downarrow$ & $\boldsymbol{\delta}_1$$\uparrow$ & \textbf{AbsRel}$\downarrow$ & $\boldsymbol{\delta}_1$$\uparrow$ \\
            \midrule
            Marigold~\cite{marigold} & 0.532 & 0.515 & 0.166 & 0.769 & 0.149 & 0.796 & 0.091 & \underline{0.931} \\
            Depth-Anything-V2~\cite{depth_anything_v2} & 0.367 & 0.554 & \underline{0.135} & 0.822 & 0.140 & 0.804 & 0.106 & 0.921 \\
            \midrule
            DepthCrafter W/O CFG & \textbf{0.270} & \textbf{0.697} & \textbf{0.123} & \textbf{0.856} & \textbf{0.104} & \textbf{0.896} & \textbf{0.071} & \textbf{0.972} \\   
            DepthCrafter W/ CFG & \underline{0.315} & \underline{0.692} & \textbf{0.123} & \underline{0.850} & \underline{0.108} & \underline{0.885} & \underline{0.076} & \textbf{0.972} \\   
            \bottomrule
        \end{tabular}
    }
    \label{supptab:CFG_scale}
\end{table*}

\section{Additional Evaluations}
\label{suppsec:quantitative}

\subsection{Effect of Number of Denoising Steps}
\label{suppsec:denoising_steps}
During inference, the number of denoising steps is a crucial hyperparameter that affects the trade-off between the inference speed and the depth estimation quality.
The practice in image-to-video diffusion models~\cite{blattmann2023stable} is to set the number of denoising steps to around 25.
However, as shown in~\cref{suppfig:steps}, we find that the number of denoising steps can be reduced significantly for video depth estimation, even one step works well.
This is because the video depth estimation task is more deterministic than the video generation task.
And we can see in the figure that more denoising steps would consistently improve the structure details of the generated depth sequences.
In~\cref{supptab:zeroshot_mde}, we show the results of our DepthCrafter with different numbers of denoising steps.
We can see that our DepthCrafter significantly outperforms existing strong baselines, such as Marigold~\cite{marigold} and Depth-Anything-V2~\cite{depth_anything_v2}, even with only one denoising step.
The performance of our DepthCrafter is increased with more denoising steps, but the improvement gets saturated after five steps.
Thus we set the number of denoising steps to five in our experiments, which achieves a good trade-off between the inference speed and the depth estimation quality.
The inference speed of our DepthCrafter with five denoising steps is 465.84 ms per 1024$\times$576 frame, which is acceptable for many applications.

\subsection{Effectiveness of Training Stages}
\label{suppsec:training_stages}
In the main paper, we ablate the performance of our DepthCrafter with three training stages, only on the Sintel~\cite{Sintel} dataset.
To complement the evaluation, we further evaluate the effectiveness of our three-stage training strategy on all the datasets, including Sintel~\cite{Sintel}, Scannet~\cite{dai2017scannet}, KITTI~\cite{Geiger2013IJRR}, and Bonn~\cite{palazzolo2019iros}.
As shown in~\cref{supptab:training_stages}, we can observe that, even only with the first two stages, our DepthCrafter already outperforms the existing strong baselines, such as Marigold~\cite{marigold} and Depth-Anything-V2~\cite{depth_anything_v2}.
More importantly, the performance improvement with the training stages is consistent across all the datasets.
It indicates that our three-stage training strategy is effective for improving the generalization ability of our DepthCrafter to diverse open-world videos.

\subsection{Effects of Classifier-Free Guidance}
\label{suppsec:classifier_free}
Classifier-free guidance (CFG) is proven to be effective in improving the details of the generated videos in video diffusion models~\cite{blattmann2023stable, chen2023videocrafter1, chen2024videocrafter2, xing2023dynamicrafter, xing2024tooncrafter}.
In our DepthCrafter, we also investigate the effectiveness of CFG in video depth estimation.
As shown in~\cref{suppfig:CFG_scale}, we show an example frame from the KITTI dataset, where the results of our DepthCrafter with and without CFG are compared.
We can see that the CFG can indeed improve the visual details of the generated depth sequences, especially for the fine-grained structures.
However, we find that the CFG may slightly degrade the quantitative accuracy of the depth estimation, as shown in~\cref{supptab:CFG_scale}.
This may be because the CFG is designed for improving the details of the generated videos, while the depth estimation task is more deterministic and requires more accurate predictions.
Since adopting the CFG would also introduce additional computation, we do not use the CFG in our DepthCrafter for the main experiments.
However, if the users are more interested in the visual details of the depth sequences, they can consider incorporating the CFG into our DepthCrafter.





\end{document}